%% file: ai2_paper.tex
\documentclass[table]{ai2style/ai2}

\usepackage{microtype}
\usepackage{hyperref}
\usepackage{url}
\usepackage{booktabs, tabularx}
\usepackage{graphicx}
\usepackage{float}
\usepackage{enumitem}
\usepackage{listings}
\usepackage{svg}

\definecolor{darkblue}{rgb}{0, 0, 0.5}
\hypersetup{colorlinks=true, citecolor=darkblue, linkcolor=darkblue, urlcolor=darkblue}

\usepackage{amssymb}
\usepackage{multirow}
\usepackage{todonotes}
\usepackage{wrapfig}
\usepackage[most]{tcolorbox}
\usepackage{xspace}
\usepackage[absolute]{textpos}

\usepackage[utf8]{inputenc}
\usepackage[T1]{fontenc}
\usepackage{amsfonts}
\usepackage[table,xcdraw]{xcolor}
\usepackage{amsmath}

\usepackage{adjustbox}
\usepackage{pifont}
\usepackage{caption}
\usepackage{makecell}

\setlist[itemize]{leftmargin=*,itemsep=0em,parsep=0.3em,topsep=0.3em}

\DeclareUnicodeCharacter{2212}{\ensuremath{-}}

\definecolor{maroon}{HTML}{F26035}
\definecolor{yellow}{HTML}{FDBC42}
\definecolor{lavender}{HTML}{734f96}
\definecolor{darkergrey}{HTML}{444444}
\definecolor{midgrey}{HTML}{e6eded}
\definecolor{sera-pink}{HTML}{F8559D}
\definecolor{sera-dark-teal}{HTML}{105257}
\definecolor{sera-bright-teal}{HTML}{0FCBBC}
\definecolor{ai2pink}{HTML}{F8559D}
\definecolor{ai2midpink}{HTML}{fad3e5}
\definecolor{ai2lightpink}{HTML}{fbecf3}
\definecolor{ai2midwhite}{HTML}{f2e5d9}
\definecolor{ai2offwhite}{HTML}{fbf4ee}
\definecolor{ai2green}{HTML}{0FCBBC}
\definecolor{ai2lightgreen}{HTML}{e7f9f3}
\definecolor{ai2darkgreen}{HTML}{105257}
\definecolor{ai2purple}{HTML}{B932EB}
\definecolor{ai2lightpurple}{HTML}{f7e8fc}
\definecolor{neutralEight}{HTML}{343434}
\definecolor{neutralFive}{HTML}{838383}
\definecolor{neutralThree}{HTML}{bebebe}
\definecolor{neutralOne}{HTML}{dedede}
\definecolor{lightgrey}{HTML}{fafcfc}
\definecolor{plum}{rgb}{0.56,0.27,0.52}

\definecolor{codebaseblue}{RGB}{31, 90, 153}
\definecolor{funcgreen}{RGB}{34, 139, 34}
\definecolor{bugorange}{RGB}{210, 105, 30}
\definecolor{modelpurple}{RGB}{128, 0, 128}
\definecolor{trajred}{RGB}{180, 50, 50}
\definecolor{patchgold}{RGB}{184, 134, 11}
\definecolor{issuecyan}{RGB}{0, 139, 139}
\definecolor{dsbrown}{RGB}{142, 112, 64}

\newcommand{\allenAiAff}{\raisebox{.28em}{\hspace{.02em}\scalebox{0.7}{\textbf{1}}}}
\newcommand{\uwAff}{\raisebox{.28em}{\hspace{.02em}\scalebox{0.7}{\textbf{2}}}}
\newcommand{\cmuAff}{\raisebox{.28em}{\hspace{.02em}\scalebox{0.7}{\textbf{3}}}}
\newcommand{\commaAff}{\raisebox{.28em}{\hspace{.02em}\scalebox{0.7}{\textbf{,}\hspace{0.1em}}}}

\newcommand{\hfdataset}{\raisebox{-1.5pt}{\includegraphics[height=1.05em]{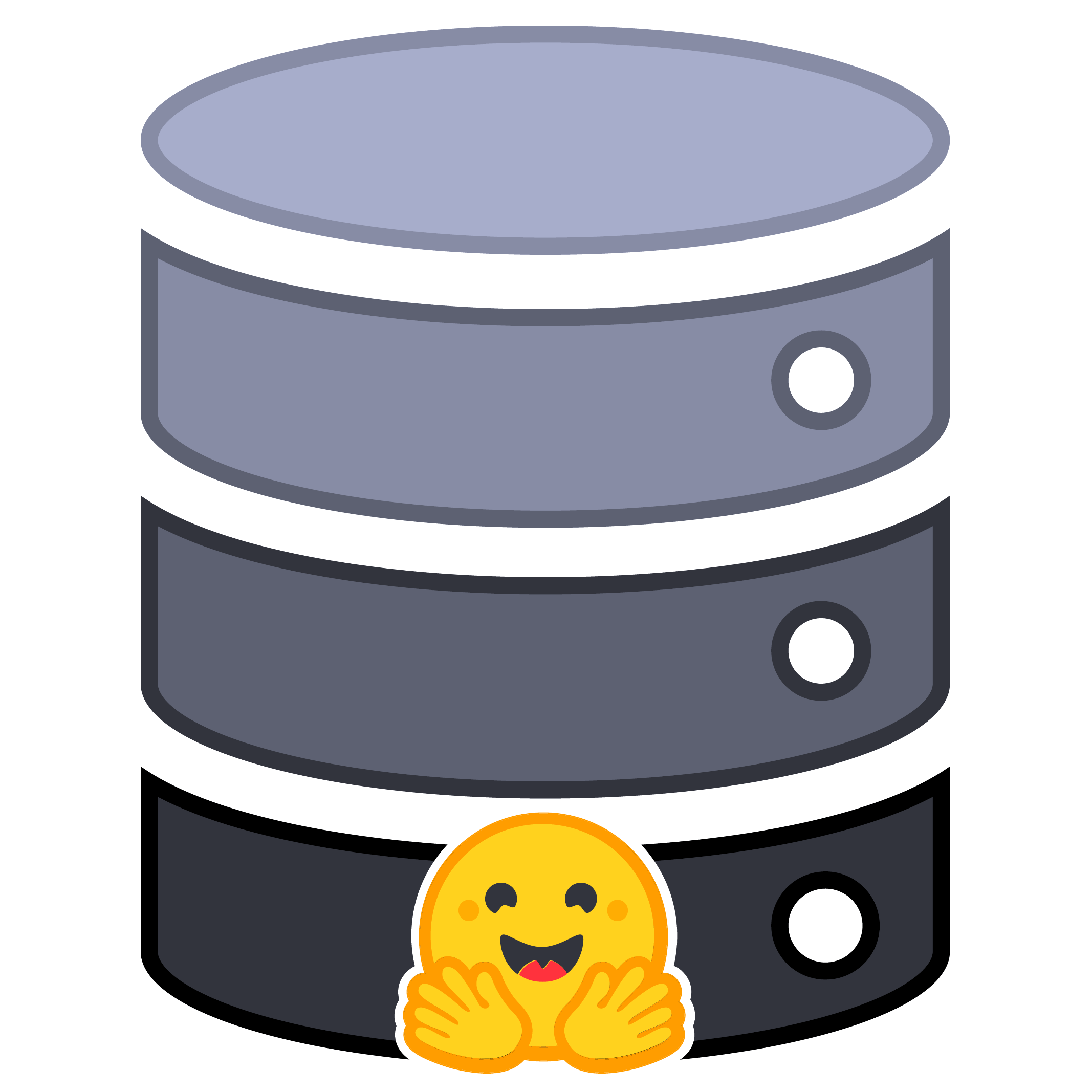}}\xspace}
\newcommand{\emailLogo}{\raisebox{-1.5pt}{\includegraphics[height=1.05em]{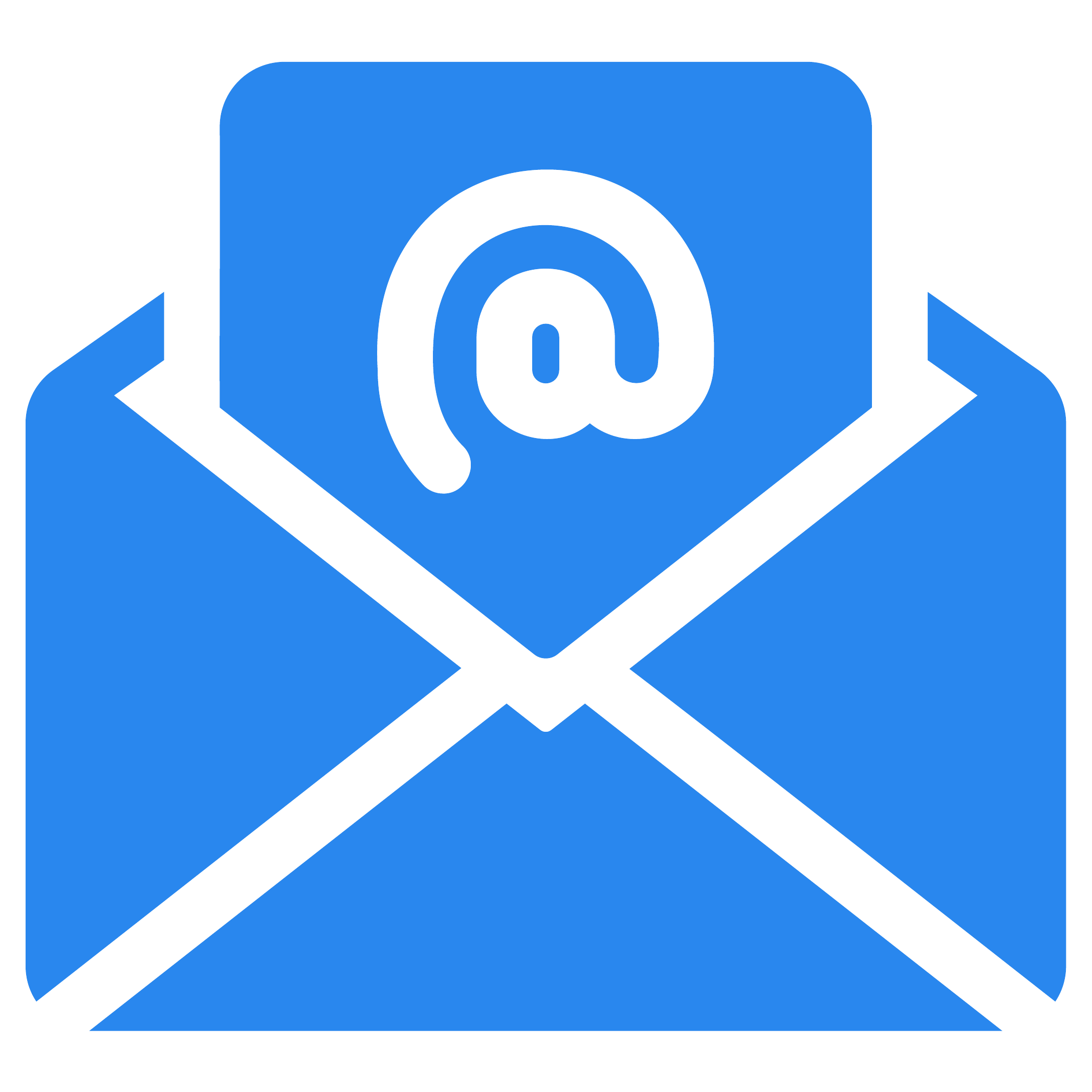}}\xspace}
\newcommand{\github}{\raisebox{-1.5pt}{\includegraphics[height=1.05em]{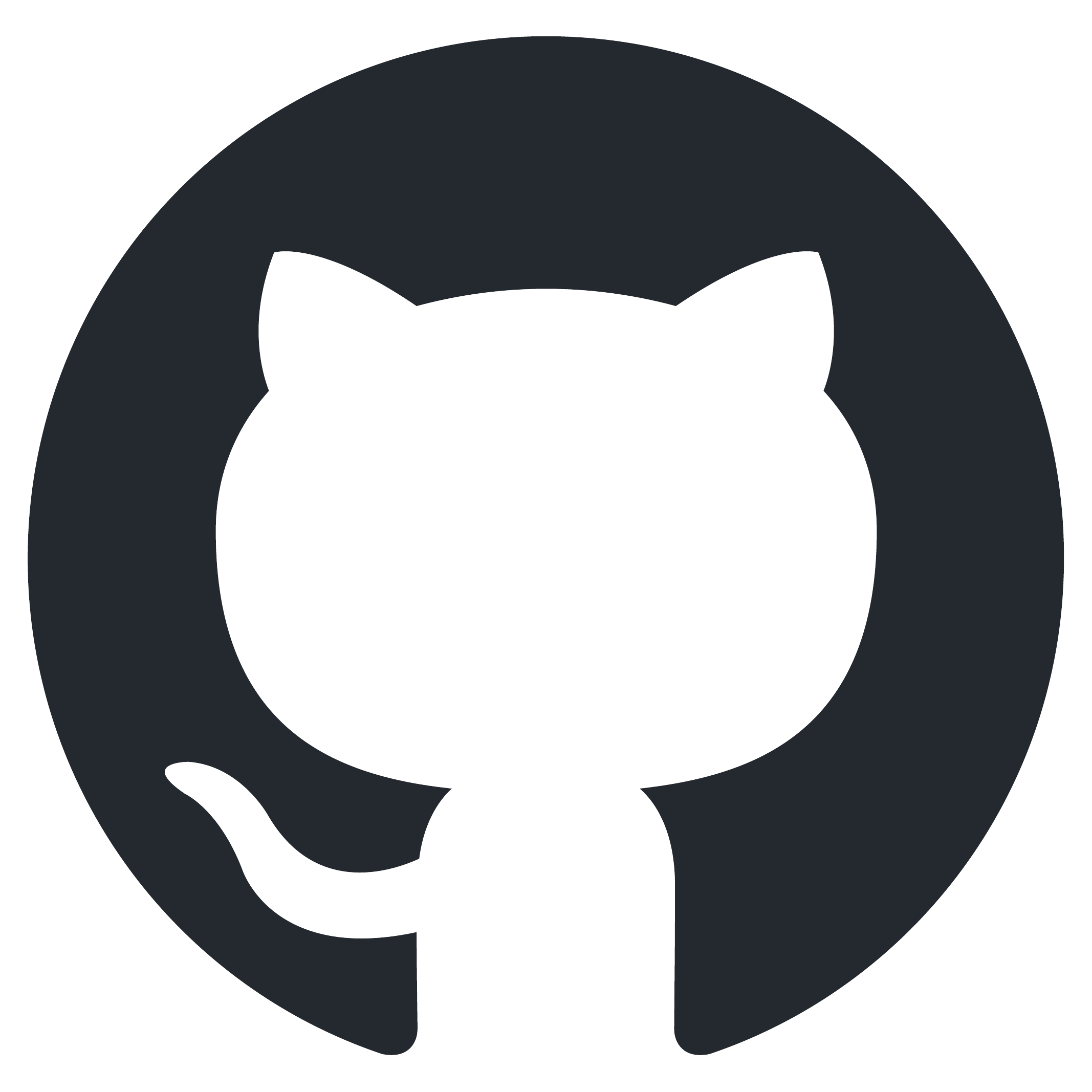}}\xspace}

\newcommand{\sera}{\textbf{\textcolor{sera-pink}{SERA}}\xspace}
\newcommand{\svg}{\textbf{\textcolor{sera-pink}{SVG}}\xspace}
\newcommand{\rollouttwo}{{\color{trajred}$\mathbf{T_2}$}\xspace}
\newcommand{\rolloutone}{{\color{trajred}$\mathbf{T_1}$}\xspace}
\newcommand{\glmairdsxone}{\textbf{\textcolor{dsbrown}{Sera-4.5A-Lite}}\xspace}
\newcommand{\glmairdsxthree}{\textbf{\textcolor{dsbrown}{Sera-4.5A-Full}}\xspace}
\newcommand{\glmdsxone}{\textbf{\textcolor{dsbrown}{Sera-4.6-Lite}}\xspace}

\title{SERA: Soft-Verified Efficient Repository Agents}

\authorOne[\allenAiAff\commaAff\uwAff]{Ethan Shen}
\authorOne[\allenAiAff]{Daniel Tormoen}
\authorOne[\allenAiAff]{Saurabh Shah}
\authorOne[\allenAiAff\commaAff\uwAff]{Ali Farhadi}
\authorOne[\allenAiAff\commaAff\cmuAff]{Tim Dettmers}

\affiliation[\allenAiAff]{Allen Institute for AI}
\affiliation[\uwAff]{University of Washington}
\affiliation[\cmuAff]{Carnegie Mellon University}

\abstract{
Open-weight coding agents should hold a fundamental advantage over closed-source systems: they can be specialized to private codebases, encoding repository-specific information directly in their weights. Yet the cost and complexity of training has kept this advantage theoretical. We show it is now practical. We present \textbf{\textcolor{sera-pink}{Soft-Verified Efficient Repository Agents (SERA)}}, an efficient method for training coding agents that enables the rapid and cheap creation of agents specialized to private codebases. Using \textbf{\textcolor{sera-pink}{Soft-Verified Generation (SVG)}}, we generate thousands of trajectories from any code repository without requiring unit tests---enabling specialization to any downstream codebase. Beyond repository specialization, we apply \svg to a larger corpus of codebases, generating 200,000+ synthetic trajectories. Using only supervised finetuning (SFT), \sera achieves state-of-the-art results among fully open-source (open data, method, code) models while matching the performance of open-weight models like Devstral-Small-2. Creating \sera models is 26x cheaper than reinforcement learning and 57x cheaper than previous synthetic data methods to reach equivalent performance. \svg is built on two observations that emerged from simplification of previous methods: First, soft verification, where instead of testing the correctness of synthetic coding data via unit tests, we only compare the partial line-by-line overlap of patches generated from two rollouts. This removes the need for test infrastructure and enables data generation from any repository, practically removing limits on the amount of data we can generate from a single codebase as well as what codebases can be used. Second, vague instructions can diversify training data, increasing the proportion of data focused on non-bug related changes like refactoring. We find that these vague instructions improve SWE-bench performance as well as bug-focused data. In more detail, \svg is based on two rollouts from an agent: in the first, a teacher model is prompted with a vague instruction to make a change to a codebase starting from a randomly selected function, producing a trajectory and patch. This trajectory is converted into a synthetic pull request. In the second, the teacher model attempts to reproduce the patch given only the pull request description. Soft verification compares the two patches using line-level recall for training data selection. Taken together, this creates a cheap pipeline for high-quality data that enables rapid experimentation. We show through power scaling curves that private codebase specialization is highly sample efficient and matches or exceeds teacher model performance at low costs. Finally, we use our data to provide detailed analysis of scaling laws, ablations, and confounding factors for training coding agents. Overall, we believe our work will greatly accelerate research on open coding agents and showcase the advantage of open-source models that can specialize to private codebases. We release \sera as the first model in Ai2's Open Coding Agents series, along with all our code, data, and Claude Code integration to support the research community.}
\metadata[\vspace{.5em}\quad\github Code:]{\color{ai2accent}\texttt{https://github.com/allenai/SERA}\quad\texttt{https://github.com/allenai/sera-cli}}
\metadata[\vspace{.5em}\quad\hfdataset Models \& Data:]{\color{ai2accent}\texttt{https://huggingface.co/collections/allenai/open-coding-agents}}
\metadata[\vspace{.5em}\quad\emailLogo Contact:]{\color{ai2accent}{\texttt{ethans03@cs.washington.edu}} \quad \color{ai2accent}{\texttt{dettmers@cmu.edu}}}

\begin{document}

\maketitle
\begin{figure}[t]
\centering
\includegraphics[width=\textwidth]{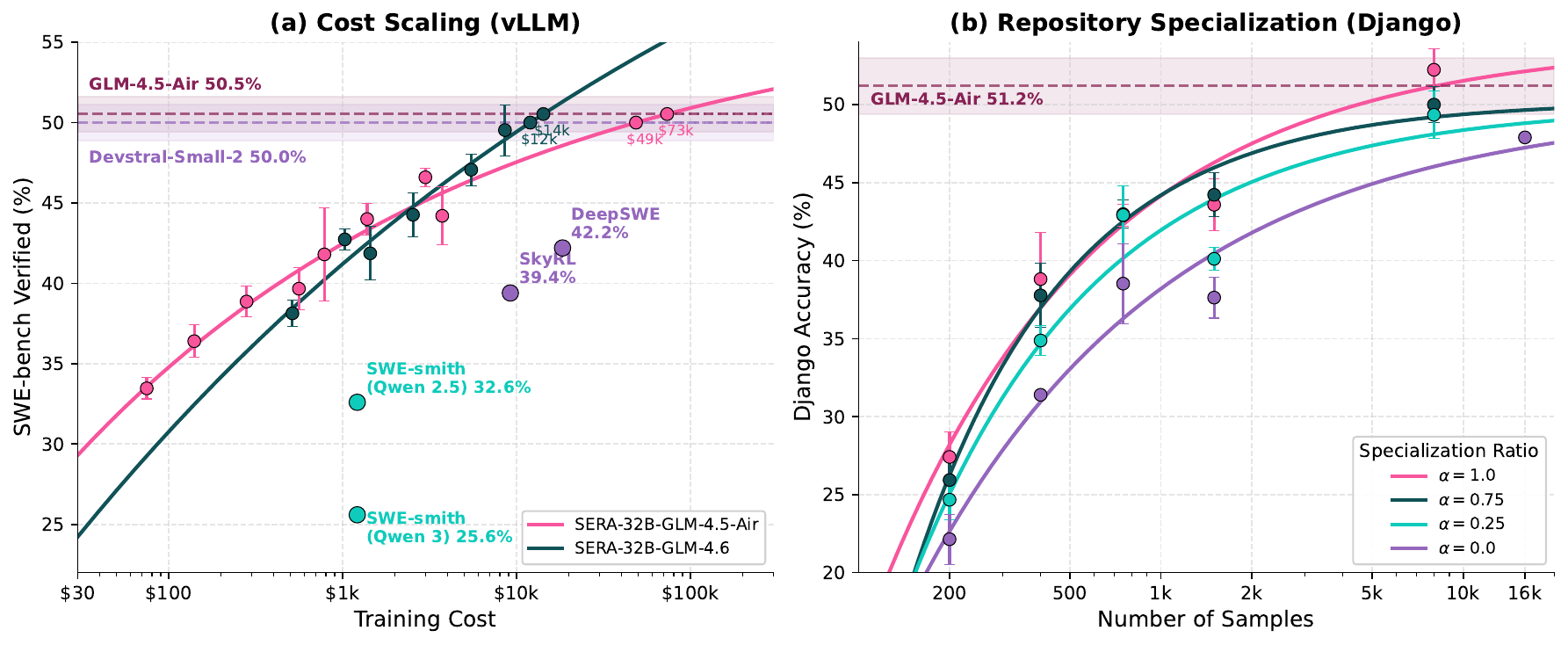}
\caption{\textbf{(a)} Scaling and cost comparison of coding agent training approaches using self-hosted vLLM inference. \textbf{(b)} Repository specialization scaling law on Django, where $\alpha$ denotes the fraction of Django-specific data in the training mixture. With full specialization ($\alpha = 1.0$), the model matches teacher performance at 8k samples; general data alone ($\alpha = 0.0$) is unable to match teacher performance, even with twice the sample size.}
\end{figure}

\begin{figure}[t]
\centering
\includegraphics[width=\textwidth]{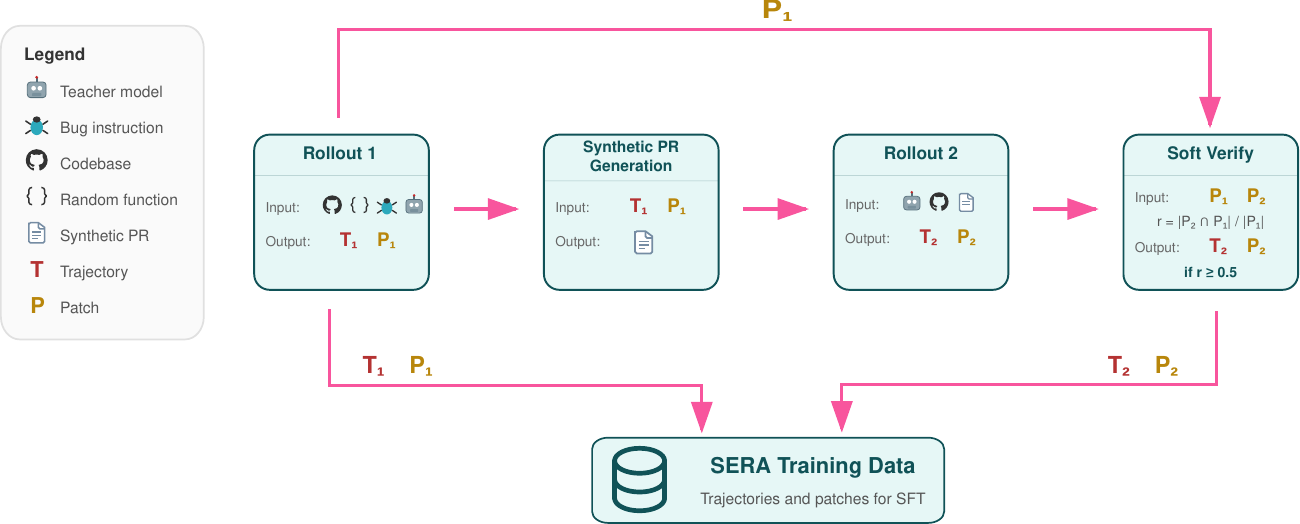}
\caption{\textbf{Overview of \svg (Soft Verified Generation).} In the first rollout, a teacher model is prompted to make a change starting from a randomly selected function, producing a trajectory and patch. This trajectory is converted into a synthetic pull request. In the second rollout, the teacher attempts to reproduce the patch given only the PR description. Soft verification compares the two patches using line-level recall for training data selection. We use $r \geq 0.5$ as an example threshold.}
\label{fig:method}
\end{figure}

\section{Introduction}
\label{sec:intro}

Coding agents have become central to software development and are increasingly applied to tasks beyond traditional engineering. While, closed-source coding agents are more powerful, open-weight models should hold a fundamental advantage in many applications because they can be specialized to private codebases, allowing them to learn repository-specific patterns, conventions, and domain knowledge. Despite this clear opportunity, the cost and complexity of training open-weight coding agents has kept this advantage theoretical. In this work, we show it is now practical. 

As the first release in Ai2's Open Coding Agents series, our method trains a 32B coding agent with simple supervised finetuning achieving state-of-the-art open-source results at 40 GPU days (\$2,000) or matching strong open-weight models like Devstral-Small-2 at a budget of \$9,000. When specializing to a particular codebase, our pipeline can match or exceed teacher model performance at \$1300.

Training coding agents traditionally requires either reinforcement learning or complex synthetic data pipelines, both demanding resources beyond what most teams can provide. Reinforcement learning requires sandboxed execution environments, distributed training infrastructure, and rollout orchestration. The complexity of this infrastructure is reflected in team sizes that average 12 or more authors in recent work \cite{cao2025skyrl, luo2025deepswe, wei2025swerl, da2025agentrlvr}. Synthetic data approaches like SWE-smith \citep{yang2025swesmith} require setting up test environments, generating valid bugs, and verifying bugs through test suites. These barriers have concentrated coding agent development in well-resourced industry labs and larger teams at academic institutions. Starting from limited compute and a small team (32 GPUs, 3 researchers), we prioritized reducing experimentation costs, which led us to systematically strip away pipeline complexity and discover which components actually matter for training effective coding agents.

We found that much of the complexity in prior pipelines is unnecessary. Firstly, soft verification, where patches are checked by partial line-by-line matching rather than executed test suites, produces training data of equal quality to full test-based verification. At the scales we test, the degree of verification has minimal effect on downstream performance, removing the need for test infrastructure entirely and enabling data generation from any repository. This makes synthetic data generation much more straightforward. Secondly, some coding instructions are inherently vague, and we observe that models prompted to fix these issues often produce changes such as refactoring or documentation improvements that are more representative of real world tasks than just bug fixes. Rather than requiring bug-focused data, we find that this general coding data is equally effective to improve performance and can be produced by prompting a model on any repository in its initial state. Together, these findings mean that generating effective training data requires neither test infrastructure nor complex bug-injection pipelines. Additionally, unlike other methods, how much data we can generate from repositories is not limited by test coverage or quality. 

The resulting cost reduction and data abundance make repository specialization practical. We show that open-weight models specialized to a codebase can match or exceed the performance of the teacher model used to generate their training data. This is intuitive: a student model with repository-specific knowledge encoded in its weights can outperform a teacher that accesses the same codebase only through its context window. The advantage this creates extends beyond privacy. Even organizations willing to share their code would need to wait many months until the next training run of a frontier model includes their data. And while LoRA adapter options for frontier models exist these are often to impractical or costly for large-scale deployments. Open-weight specialization allows practitioners to generate data from their repositories, fine-tune, and deploy immediately, iterating as their codebase evolves. At low cost, any team can build and deploy a small specialized model that outperforms frontier systems on their own codebase -- an advantage that grows with the codebase and that frontier labs cannot close regardless of their scale.

We introduce \sera (Soft-verified Efficient Repository Agents), a 32B coding agent that achieves 49.5\%/54.2\% on SWE-bench Verified when evaluated at 32K/64K context, state-of-the-art for fully open-source models. We exceed the performance of previous open-source solutions with a total cost of \$2,000 for both data generation and training (40 GPU days). Our method, \svg (Soft Verified Generation; see Figure~\ref{fig:method} for an overview), achieves equivalent performance to SkyRL at 26$\times$ lower cost and to SWE-smith at 57$\times$ lower cost when self-hosting inference via vLLM. Using the z.ai API, these advantages increase to 53$\times$ and 115$\times$ respectively. These efficiency factors are derived from our scaling laws that capture both per-sample savings and data quality gains (see Appendix~\ref{sec:cost} for detailed cost breakdowns). Effectively specializing to a single repository requires approximately 8,000 trajectories (\$1,300). These trajectories are generated from randomly selected functions and contain no information about evaluation issues or their solutions. We validate our findings across multiple seeds and use scaling laws as robustness checks, adopting a methodology designed to ensure reported effects reflect genuine signal rather than noise. To further support open coding agent research, we provide extensive analyses covering ablations on data quality factors, model-specific pitfalls, and common confounding factors that have slowed progress in this area.

We release \sera as the first model in Ai2's Open Coding Agents series, along with all our code, 200,000 synthetic coding agent trajectories, and Claude Code integration.

\section{Background}
\label{sec:background}

This section introduces the core concepts and prior approaches for training coding agents. We cover the standard evaluation benchmark, the structure of agent systems and their training data, and the two main paradigms for training on synthetic data: data generation and reinforcement learning.

\subsection{SWE-bench}

SWE-bench \citep{jimenez2024swebench} is the standard benchmark for evaluating coding agents on real-world software engineering tasks. While other benchmarks exist that are more comprehensive, such as Terminal-Bench \citep{merrill2026terminalbench}, SWE-bench offers broader comparisons between methods and models, and its confounding factors are better understood.
Each task is derived from a real GitHub issue and pull request from $12$ popular Python repositories such as Django, Sympy, and Sphinx. Given an issue description, the agent must produce a patch that resolves the issue. The repository's test suite is run before and after applying the patch and a task is considered solved if previously failing tests now pass and no previously passing tests are broken. SWE-bench Verified is a curated subset where human annotators have verified that each task is solvable and that the tests correctly validate the solution. We use SWE-bench Verified for all evaluations.

\subsection{Agent Scaffolds and Training Data}

Coding agents operate through scaffolds that define the tools available to the agent and how it interacts with the environment. SWE-agent \citep{yang2024sweagent} is a widely used scaffold that provides tools for viewing files, editing code, and executing bash commands. The agent receives observations from these tools and produces actions in a loop until the agent decides to perform the last action, which in this case of SWE-bench is the submission of the final software patch.

A \textbf{rollout} is one complete execution of the agent on a task, from receiving the issue to submitting a solution. The sequence of actions, observations, and reasoning produced during a rollout is called a \textbf{trajectory}. A \textbf{patch} is the final output: a line-by-line diff specifying additions and deletions to the codebase.

Trajectories are the training data for coding agents. The two main approaches for generating trajectories are synthetic data generation and reinforcement learning, which we describe in the following subsections.

Two practical considerations affect both approaches. First, coding agent trajectories can be very long, often requiring 32K tokens or more of context \citep{yang2024sweagent}. This makes experimentation slow and expensive, and means that models must handle long contexts well and efficiently to be effective coding agents. Second, coding agents rely heavily on tool calling: if a model cannot reliably follow the tool format and produce valid tool calls, it cannot function as a coding agent regardless of its other capabilities.

\subsection{Synthetic Data Generation}

Synthetic data generation creates trajectories by having a strong teacher model solve synthetic tasks---artificially constructed problems designed to mimic real issues but generated programmatically rather than from actual user reports---then using those trajectories to train a smaller student model. This teacher-student distillation approach separates data generation from training, allowing each to be optimized independently.

The standard approach, exemplified by SWE-smith \citep{yang2025swesmith}, generates training data through bug injection. Starting with a repository that has a passing test suite, the pipeline programmatically injects bugs that cause tests to fail, generates an issue description from the bug, has the teacher model solve the issue, and verifies correctness by checking that tests pass again. This requires test infrastructure, valid bug generation, and execution environments for verification.

The cost structure of this approach is significant. Using Sonnet 3.7, each SWE-smith trajectory costs approximately \$0.52 including issue creation and rollout (see Section~\ref{sec:cost} for a detailed cost comparison).
Beyond direct costs, the complexity of CPU-based test execution adds complex infrastructure and slows iteration on experiments.

A limitation of teacher-student distillation is that the student's performance is largely bounded by the teacher's capability. While there are cases where students can slightly exceed their teachers \citep{hinton2015distilling, furlanello2018born}---and we see similar results in our work (see Sections~\ref{sec:scaling} and~\ref{sec:specialize})---the gains are modest.
This means that at the frontier, where no stronger teacher exists, synthetic data generation may not be sufficient for further progress and reinforcement learning might be required.

\subsection{Reinforcement Learning}

Reinforcement learning trains coding agents by having them generate rollouts and learning from reward signals based on whether tasks are solved. Unlike synthetic data generation, the model being trained is also the model generating trajectories.

This has both advantages and disadvantages. The disadvantage is that if the model is initially too weak, improvement is slow or nonexistent because it generates mostly unsuccessful trajectories to learn from. The advantage is that a strong model can continue to improve through self-play, since it is not bounded by a separate teacher's capabilities. At the very frontier of model performance, reinforcement learning may be the only path to further progress.

However, reinforcement learning introduces substantial infrastructure complexity. It requires online rollouts during training, sandboxed execution environments, and distributed systems for coordinating rollouts with gradient updates. This complexity is reflected in team sizes: recent RL papers for software engineering agents average 12 or more authors, including SWE-RL with 9 authors \citep{wei2025swerl}, SkyRL-Agent with 15 authors \citep{cao2025skyrl}, and SWE-rebench with 9 authors \citep{badertdinov2025swerebench}. Beyond team size, reinforcement learning is known to be highly unstable and difficult to use in practice, with training sensitive to hyperparameters, reward shaping, and random seeds \citep{henderson2018deep, engstrom2020implementation}.

For small academic teams like ours, with one researcher, two engineers, and two advisors, reinforcement learning approaches are difficult to execute. This motivated our focus on supervised methods that achieve comparable results with simpler infrastructure.

\subsection{Verification}

Verification determines whether a generated trajectory is suitable for training. Traditional approaches use \textbf{unit test verification}: the patch must pass all relevant tests, confirming that the synthetic bug was correctly resolved. This ensures correctness but limits data generation to repositories with comprehensive test coverage and sufficient test quality.

Our method introduces \textbf{soft verification}: instead of executing tests, we compare the generated patch against a reference patch using line-level recall. If the generated patch contains most or all of the lines from the reference patch, we consider it verified. This removes the need for test infrastructure and enables data generation from any repository. We describe the details of our soft verification approach in the following section.

\section{Method}
\label{sec:method}

\subsection{{\color{sera-pink}Soft Verified Generation} (\svg)}

The key intuition behind \svg is that clear errors in syntax, logic, and failing unit tests are only a subset of real world coding pull requests (PRs). Indeed, it is extremely common for PRs to be more obscure, aimed at refactoring code, enforcing style requirements, or tweaking behavior. This section will be an overview of \svg and then describe each of the components in more detail. Figure~\ref{fig:method} visualizes every step of \svg.

In \svg, we rethink the criteria that define a valid synthetic PR. While traditional synthetic approaches explicitly focus synthetic issues on failed unit tests to ensure samples represent correct code, we instead broaden the definition of a PR to include any instruction that attempts to create \textit{some} desired change in a codebase {\color{codebaseblue}$\mathbf{C}$}. This interpretation is central to our approach. The key insight is that a trajectory's value for training lies not in producing a fully correct patch, but in the skills it demonstrates, for example, how to interpret an instruction, navigate a codebase, and translate intent into code.

\svg is composed of two rollouts. We use {\color{trajred}$\mathbf{T}$} and {\color{patchgold}$\mathbf{P}$} to denote the trajectory and patch created by a rollout. In \svg, we use a teacher model {\color{modelpurple}$\mathbf{M}$} to generate rollouts. In the first rollout, we prompt {\color{modelpurple}$\mathbf{M}$} with a random function {\color{funcgreen}$\mathbf{func_i}$} from codebase {\color{codebaseblue}$\mathbf{C}$} and a bug prompt {\color{bugorange}$\mathbf{bug_j}$} sampled from a set of 51 bug types {\color{bugorange}$\mathbf{B}$}. This produces trajectory {\color{trajred}$\mathbf{T_1}$} and patch {\color{patchgold}$\mathbf{P_1}$}. We then convert {\color{trajred}$\mathbf{T_1}$} into a synthetic PR {\color{issuecyan}$\mathbf{synth\_PR}$} using a demonstration PR {\color{issuecyan}$\mathbf{PR}$} sampled from SWE-Bench Verified. In the second rollout, {\color{modelpurple}$\mathbf{M}$} is prompted with {\color{issuecyan}$\mathbf{synth\_PR}$} and tasked to reproduce the original change, producing trajectory {\color{trajred}$\mathbf{T_2}$} and patch {\color{patchgold}$\mathbf{P_2}$}. Soft-verification compares {\color{patchgold}$\mathbf{P_2}$} against {\color{patchgold}$\mathbf{P_1}$} using line-level recall $r$. The combination of these steps is \svg. We provide a general mathematical overview below and more detail in the following sections.

\begin{equation}
\llap{\textbf{First Rollout:}\quad} {\color{trajred}\mathbf{T_1}}, {\color{patchgold}\mathbf{P_1}} = {\color{modelpurple}\mathbf{M}}({\color{funcgreen}\mathbf{func_i}}, {\color{bugorange}\mathbf{bug_j}}, {\color{codebaseblue}\mathbf{C}})
\end{equation}
\begin{equation}
\llap{\textbf{Synthetic PR Generation:}\quad} {\color{issuecyan}\mathbf{synth\_PR}} = {\color{modelpurple}\mathbf{M}}({\color{trajred}\mathbf{T_1}}, {\color{issuecyan}\mathbf{PR}})
\end{equation}
\begin{equation}
\llap{\textbf{Second Rollout:}\quad} {\color{trajred}\mathbf{T_2}}, {\color{patchgold}\mathbf{P_2}} = {\color{modelpurple}\mathbf{M}}({\color{issuecyan}\mathbf{synth\_PR}}, {\color{codebaseblue}\mathbf{C}})
\end{equation}
\begin{equation}
\llap{\textbf{Soft Verification:}\quad} r = \frac{|{\color{patchgold}\mathbf{P_2}} \cap {\color{patchgold}\mathbf{P_1}}|}{|{\color{patchgold}\mathbf{P_1}}|}
\end{equation}
If $r = 1$, the trajectory is hard-verified; if $0 < r < 1$, soft-verified; if $r = 0$, unverified.

We will now explain each component in depth.

\textbf{Agent Workflow:} We use SWE-agent \citep{yang2024sweagent} to generate trajectory rollouts. SWE-agent allows users to define a variety of tools available to an agent and gives users the ability to adjust settings such as the length of tool outputs and context history. To reduce the effect of confounding factors, we use SWE-agent in its vanilla state: we only provide the agent with the ability to run a view tool, edit tool, submission tool, and bash commands. Furthermore, we do not truncate context history or tool outputs at any point during rollouts. While truncations are frequently done to avoid context window errors, we noticed that many previous works use slightly different heuristics, making it difficult to objectively compare performance. Additionally, we believe an important trait of coding agents is their ability to solve tasks while avoiding unnecessarily long tool calls and outputs. 

\textbf{First Rollout:} At a high-level, the first rollout works as follows: We prompt the teacher model {\color{modelpurple}$\mathbf{M}$} with ``There is a {\color{bugorange}$\mathbf{bug_j}$} related to- function {\color{funcgreen}$\mathbf{func_i}$}.'', where  {\color{bugorange}$\mathbf{bug_j}$} is a high-level description of a bug type, {\color{funcgreen}$\mathbf{func_i}$} is a randomly chosen function in the codebase {\color{codebaseblue}$\mathbf{C}$}. The function {\color{funcgreen}$\mathbf{func_i}$} serves as an arbitrary starting point for the agent. We run the pipeline once for every function in the codebase. Each {\color{bugorange}$\mathbf{bug_j}$} is randomly sampled from a larger list of 51 types of bugs {\color{bugorange}$\mathbf{B}$} and asks the model to fix issues ranging from state management to code clarity. We generate this list of 51 bugs from papers that study bug distributions in software systems \citep{just2014defects4j, widyasari2020bugsinpy}. We intentionally leave the prompt vague to widen the range of acceptable changes. We rollout for a maximum of 115 steps, although this limit is rarely reached. 

Occasionally, the model is unable to find a bug that both aligns with the prompt and is related to the provided function. To handle this edge case, we separately ask the teacher model {\color{modelpurple}$\mathbf{M}$} to self-evaluate its fix after the rollout is finished. We accept the trajectory {\color{trajred}$\mathbf{T_1}$} unless the teacher model decides it did not make a change aligned with the prompt. In that case, we reject the trajectory and we perform another rollout with a different sampled prompt until a valid change is produced or a limit of three runs is reached. About 2\% of rollouts are rejected during the first rollout, and less than 1\% fail all three rollouts. We also discard trajectories that produce duplicate patches, although we observe that this is extremely rare. The final patches {\color{patchgold}$\mathbf{P_1}$} from accepted trajectories are saved as ground truth. Importantly, these synthetic tasks are generated without reference to evaluation benchmarks—no information about the bugs or GitHub issues in SWE-bench Verified is contained in our training data.
\begin{equation}
{\color{trajred}\mathbf{T_1}}, {\color{patchgold}\mathbf{P_1}} = {\color{modelpurple}\mathbf{M}}({\color{funcgreen}\mathbf{func_i}}, {\color{bugorange}\mathbf{bug_j}}, {\color{codebaseblue}\mathbf{C}}),
\end{equation}

\textbf{Synthetic PR:} Next, to create {\color{issuecyan}$\mathbf{synth\_PR}$} to guide the second stage, we provide the teacher model {\color{modelpurple}$\mathbf{M}$} with its first rollout {\color{trajred}$\mathbf{T_1}$}, which contains relevant reproduction scripts, execution traces, and the final software patch. Similar to SWE-smith \citep{yang2025swesmith}, we also include a demonstration PR {\color{issuecyan}$\mathbf{PR}$} sampled from SWE-Bench Verified \citep{jimenez2024swebench}. The teacher is then asked to write a new PR that follows the format of the demonstration PR.
\begin{equation}
{\color{issuecyan}\mathbf{synth\_PR}} = {\color{modelpurple}\mathbf{M}}({\color{trajred}\mathbf{T_1}}, {\color{issuecyan}\mathbf{PR}}),
\end{equation}

\textbf{Second Rollout:} In the second rollout, we only use the synthetic PR {\color{issuecyan}$\mathbf{synth\_PR}$} as the input, with the goal of reproducing the initial patch. The trajectory {\color{trajred}$\mathbf{T_2}$} is again capped at 115 steps, and the resulting patch {\color{patchgold}$\mathbf{P_2}$} is saved.
\begin{equation}
{\color{trajred}\mathbf{T_2}}, {\color{patchgold}\mathbf{P_2}} = {\color{modelpurple}\mathbf{M}}({\color{issuecyan}\mathbf{synth\_PR}}, {\color{codebaseblue}\mathbf{C}}),
\end{equation}

\textbf{Soft Verification:} We evaluate the second rollout patch {\color{patchgold}$\mathbf{P_2}$} using recall against the first rollout patch {\color{patchgold}$\mathbf{P_1}$} by assessing edits at a line-by-line granularity. If {\color{patchgold}$\mathbf{P_2}$} contains every change from {\color{patchgold}$\mathbf{P_1}$}, then the recall is $r = 1$ and the second rollout is considered hard-verified. If $0 < r < 1$, then the rollout is considered soft-verified. Finally, if $r=0$, then it is considered unverified.
\begin{equation}
r = \frac{|{\color{patchgold}\mathbf{P_2}} \cap {\color{patchgold}\mathbf{P_1}}|}{|{\color{patchgold}\mathbf{P_1}}|},
\end{equation}
where $r$ is the line-level recall.

\textbf{Setup Details:} We use a suite of 121 codebases for data generation, which are a subset of the 128 codebases released by SWE-smith \citep{yang2025swesmith}.\footnote{We exclude 7 codebases that contain little to no Python code, such as repositories consisting of a single file or minimal Python content.} Each codebase {\color{codebaseblue}$\mathbf{C}$} is encapsulated inside of a docker container. We use GLM-4.5-Air \citep{zeng2025glm45} as our teacher model {\color{modelpurple}$\mathbf{M}$} for all experiments unless otherwise specified. GLM-4.5-Air has several advantages: it allows for scaled experiments, has full reasoning traces, is powerful, and is easy to deploy on commonly available GPUs. Pairing \svg's generation efficiency and GLM-4.5-Air's cost efficiency makes a robust scientific investigation of coding agent scaling possible.

Indeed, current data generation strategies for coding agents are often bottlenecked by their reliance on closed-source models as teachers, whose API costs make studies of coding agent scaling impractical and hamper statistical reliability, since repeating evaluations across different random seeds becomes very costly. Furthermore, closed-source providers often hide full reasoning traces, which are essential for data quality (Section \ref{sec:teachers}) and APIs are prone to change or adjust model quality depending on demand. As an open-weight model, GLM-4.5-Air can be run locally, avoiding these issues. GLM-4.5-Air also strikes a powerful balance between performance and model size: it provides Claude 3.7 Sonnet\footnote{\url{https://www.anthropic.com/news/claude-3-7-sonnet}} level performance while being fully deployable on 8 H100s, or 4 H100s at a lower context length, or 2 H100s if quantized. We hope that this will significantly reduce barriers for practitioners and researchers who want to train, use, and study coding agents at scale.

\subsection{Training}
We use Qwen 3-32B \citep{yang2025qwen3} as our primary base model over models like Qwen 2.5 \citep{yang2024qwen25} due to Qwen 3-32B's stronger tool calling performance, which better reflects the improving capabilities of current and future base models. This mirrors similar choices from recent work on coding agents \citep{sonwane2025bugpilotcomplexbuggeneration, cao2025skyrl, luo2025deepswe}. 
We fully fine-tune up to Qwen 3's native context length of $32768$ and train our models for 3 epochs using a learning rate of \texttt{1e-5} and weight decay of $0.01$. We primarily use axolotl \citep{axolotl} for training and vLLM \citep{kwon2023vllm} for model hosting.

We prioritize training on trajectories that are $\leq$32K tokens in length. To increase sample size as needed, we selectively truncate longer trajectories based on the ratio of trajectory steps within the context limit---we term this ``truncation ratio''. In Section \ref{sec:trunc}, we explore the effects of truncation in depth and explain why it must be done with caution.

\section{Main Results}
\label{sec:results}

\begin{table}[!htb]
\centering
\resizebox{\textwidth}{!}{%
\begin{tabular}{l|ccc|c|c|c|l}
\toprule
& \multicolumn{3}{c|}{\textbf{Open Source}} & & & & \\
\textbf{Method} & \textbf{Code} & \textbf{Model} & \textbf{Data} & \textbf{Base Model} & \textbf{Teacher} & \textbf{Context} & \textbf{Resolve Rate} \\
\midrule
SkyRL-8B & \ding{51} & \ding{51} & & Qwen 3-8B & --- & 32K & 9.4\% \\
Nex-N1-8B & \ding{51} & \ding{51} & & InternLM3-8B & --- & 32K & 20.3\% \\
\midrule
\rowcolor{pink!20} \sera-8B-GA (Ours) & \ding{51} & \ding{51} & \ding{51} & Qwen 3-8B & GLM-4.5-Air & 32K & 31.7\% $\pm$ 0.4\% \\
\rowcolor{pink!20} \sera-8B (Ours) & \ding{51} & \ding{51} & \ding{51} & Qwen 3-8B & GLM-4.6 & 32K & 31.7\% $\pm$ 0.9\% \\
\midrule
\midrule
\rowcolor{gray!20} Qwen 3-32B & \ding{55} & \ding{51} & \ding{55} & 32B & --- & 32K & 24.4\% \\
SWE-smith & & & & Qwen 3-32B & Claude 3.7 & 32K & 25.6\% $\pm$ 1.1\% \\
SWE-smith & \ding{51} & \ding{51} & \ding{51} & Qwen 2.5-32B & Claude 3.7 & 32K & 32.6\% \\
\rowcolor{gray!20} FrogBoss-32B & \ding{55} & \ding{51} & \ding{55} & Qwen 3-32B & Claude 4 Sonnet & 32K & 35.0\% \\
\rowcolor{gray!20} GLM-4.7-Flash & \ding{55} & \ding{51} & \ding{55} & 30B & --- & 32K & 37.3\% $\pm$ 2.0\% \\
SkyRL-Agent & \ding{51} & \ding{51} & & Qwen 3-32B & --- & 32K & 39.4\% \\
DeepSWE & \ding{51} & \ding{51} & & Qwen 3-32B & --- & 32K & 42.2\% \\
\rowcolor{gray!20} Qwen 3-Coder-30B & \ding{55} & \ding{51} & \ding{55} & 30B & --- & 32K & 45.0\% \\
\rowcolor{gray!20} Kimi-dev & \ding{55} & \ding{51} & \ding{55} & 72B & --- & 32K & 48.6\% \\
\rowcolor{gray!20} Devstral-Small-2 & \ding{55} & \ding{51} & \ding{55} & 24B & --- & 32K & 50.0\% $\pm$ 1.3\% \\
\rowcolor{gray!20} GLM-4.5-Air & \ding{55} & \ding{51} & \ding{55} & 110B & --- & 32K & 50.5\% $\pm$ 1.3\% \\
\rowcolor{gray!20} GLM-4.6 & \ding{55} & \ding{51} & \ding{55} & 357B & --- & 32K & 60.8\% \\
\midrule
\rowcolor{pink!20} \sera-32B-GA (Ours) & \ding{51} & \ding{51} & \ding{51} & Qwen 3-32B & GLM-4.5-Air & 32K & 46.6\% $\pm$ 0.7\% \\
\rowcolor{pink!20} \sera-32B (Ours) & \ding{51} & \ding{51} & \ding{51} & Qwen 3-32B & GLM-4.6 & 32K & 49.5\% $\pm$ 1.9\% \\
\midrule
\midrule
\rowcolor{gray!20} GLM-4.7-Flash & \ding{55} & \ding{51} & \ding{55} & 30B & --- & 64K & 39.7\% $\pm$ 1.8\% \\
SWE-Swiss & \ding{51} & \ding{51} & & Qwen 2.5-32B & --- & 128K & 45.0\% \\
\rowcolor{gray!20} Qwen 3-Coder-30B & \ding{55} & \ding{51} & \ding{55} & 30B & --- & 256K & 51.6\% \\
\rowcolor{gray!20} CWM & \ding{55} & \ding{51} & \ding{55} & 32B & --- & 128K & 53.9\% \\
\rowcolor{gray!20} FrogBoss-32B & \ding{55} & \ding{51} & \ding{55} & Qwen 3-32B & Claude 4 Sonnet & 64K & 54.6\% \\
\rowcolor{gray!20} GLM-4.5-Air & \ding{55} & \ding{51} & \ding{55} & 110B & --- & 64K & 57.4\% $\pm$ 0.5\% \\
\rowcolor{gray!20} Devstral-Small-2 & \ding{55} & \ding{51} & \ding{55} & 24B & --- & 64K & 59.1\% $\pm$ 1.1\% \\
\rowcolor{gray!20} GLM-4.7-Flash & \ding{55} & \ding{51} & \ding{55} & 30B & --- & 128K & 59.2\% \\
\rowcolor{gray!20} Devstral-Small-2 & \ding{55} & \ding{51} & \ding{55} & 24B & --- & 256K & 68.0\% \\
\midrule
\rowcolor{pink!20} \sera-32B-GA (Ours) & \ding{51} & \ding{51} & \ding{51} & Qwen 3-32B & GLM-4.5-Air & 64K & 51.7\% $\pm$ 1.1\% \\
\rowcolor{pink!20} \sera-32B (Ours) & \ding{51} & \ding{51} & \ding{51} & Qwen 3-32B & GLM-4.6 & 64K & 54.2\% $\pm$ 1.4\% \\
\bottomrule
\end{tabular}%
}
\caption{SWE-bench Verified performance comparing \sera against other coding agent training methods. We separate by sequence length as this is the largest confounding factor. Gray rows are open-weight models, white rows are fully open-source models. Standard deviations reported where available from our replications using 3 random seeds. Nex-N1-8B from \citet{cai2025nexn1}, InternLM3-8B from \citet{cai2024internlm2}, CWM from \citet{copet2025cwm}.}
\label{tab:leaderboard}
\end{table}
We primarily evaluate on SWE-bench Verified \citep{jimenez2024swebench}, a curated subset of SWE-bench where human annotators have verified that each task is solvable and that the tests correctly validate the solution. In this section, we focus on three evaluation settings: (1) a head-to-head comparison that controls for the teacher model while comparing against other synthetic data methods, (2) a scaling law study that examines how our approach scales with data size and predicts when we reach certain performance thresholds, and (3) a focused benchmarking of how well our approach can target specific codebases for improved performance.

A key consideration in our evaluation methodology is controlling for evaluation context length. Context length has a significant impact on memory footprint, even among models of equal size. Doubling context length often requires increasing memory by nearly the same factor. We also observe that context length is one of the factors that strongly differentiates model performance---methods evaluated at 64K or 128K context often appear substantially stronger than those evaluated at 32K context, even when the underlying model capabilities are similar. To ensure fair comparisons across deployment configurations, we explicitly report and control for context length in all experiments, and we group results by context size in our leaderboard (Table~\ref{tab:leaderboard}). For every experiment, performance is averaged across three random seeds, a practice we find essential for reliable conclusions given the high variance in coding agent evaluations (see Section~\ref{sec:eval_distribution} for detailed statistical analysis).
\subsection{Controlled Comparisons}
\label{sec:compare}

\begin{table}[h!]
\centering
\begin{tabular}{l|c|c||c|c}
\toprule
Method & SWE-smith & \sera & BugPilot & \sera \\
\midrule
Base model & Qwen 3-32B & Qwen 3-32B & Qwen 3-32B & Qwen 3-32B \\
Teacher & Claude 3.7 & Claude 3.7 & Claude 4 Sonnet & Claude 4 Sonnet \\
Eval context size & 32K & 32K & 64K & 64K \\
Sample size & 4776 & 4776 & 5819 & 5319 \\
\midrule
SWE-bench Verified & 25.27\% $\pm$ 0.61\% & 30.00\% $\pm$ 1.41\% & 49.87\% & 48.53\% $\pm$ 0.31\% \\
\bottomrule
\end{tabular}
\caption{SWE-Bench Verified results comparing SWE-smith baseline with \sera across different teacher models and context sizes. All experiments use Qwen 3-32B as the base model. Additional baseline comparisons including SWE-smith on Qwen 2.5 and \sera with GLM-4.5-Air are provided in Appendix~\ref{sec:additional_baselines}.}
\label{tab:swebench-verified}
\end{table}
The goal of this section is to understand the differences between \sera and other synthetic data generation methods when we control for teacher model, verification method, and evaluation context length. Because hard verification rates vary based on repository difficulty and teacher model capability, sample sizes can only be approximately matched---we generate samples and filter post-hoc for hard verification, which introduces some variance in the final dataset sizes.

We also note an important methodological consideration regarding context management during evaluation. Some agent frameworks employ optimizations such as retaining only the last few tool calls in context rather than the full trajectory history. While this compression allows models to appear effective at longer context lengths, it introduces a confounding factor: methods using such optimizations may appear to benefit from increased context without the associated computational cost. Furthermore, such optimizations cause key-value cache invalidation during inference, which is prohibitively expensive for practical deployment. For fair comparison, we evaluate all methods using full context retention without such optimizations, ensuring that reported context lengths accurately reflect the actual information available to the model.

We compare against SWE-smith and BugPilot using hard-verified trajectories from the second rollout. This ensures that our training data distribution mimics that of other synthetic setups (i.e. synthetic issue descriptions and working code). From Table~\ref{tab:swebench-verified}, in a head-to-head comparison with the same teacher and sample size, \sera yields better performance trained on hard-verified trajectories. Additional comparisons showing that SWE-smith is optimized for Qwen 2.5 and that GLM-4.5-Air provides substantial improvements as a teacher model are provided in Appendix~\ref{sec:additional_baselines}.

We also evaluate at 64K context size and compare against BugPilot BaseMix \citep{sonwane2025bugpilot}, the base mixture in BugPilot's training data, which combines real and synthetic issues from R2EGym and SWE-smith using Claude 4 Sonnet as the teacher. Because BugPilot's data is not public, we choose BaseMix because its reported sample size is closest to our largest Claude 4 Sonnet run. Still, our train set contains approximately 10\% fewer samples. Despite this, in a head-to-head comparison, our results nearly match BugPilot's performance.

These results demonstrate that the data quality of our approach is high. Even when controlling for the teacher model, \sera matches real and synthetic approaches that use complicated bug generation pipelines and unit test verification.

\subsection{Scaling Experiments}
\label{sec:scaling}

\sera significantly simplifies the process of generating massive amounts of coding data by circumventing the need to introduce synthetic bugs into codebases and validate them with unit test execution. We take advantage of this property to generate three large-scale datasets from the codebases described in Section \ref{sec:method}, using both GLM-4.5-Air and GLM-4.6 as teachers.
\begin{itemize}
    \item \glmairdsxone is generated by running our data generation pipeline once for every function across all 121 codebases using GLM-4.5-Air as the teacher. This results in approximately 36,000 \rolloutone and 36,000 \rollouttwo trajectories. 
    \item \glmairdsxthree is a superset of \glmairdsxone. We continue our generation from \glmairdsxone, looping through every {\color{funcgreen}$\mathbf{func_i}$} up to three total times. Each time, a new bug prompt is sampled for the first rollout. This ensures that every trajectory is unique even for the same {\color{funcgreen}$\mathbf{func_i}$}. We stop generation after several days, reaching a total of 70,000 \rolloutone and 70,000 \rollouttwo trajectories.
    \item \glmdsxone mimics the setup of \glmairdsxone, but uses GLM-4.6 as the teacher model. We generate another 36,000 \rolloutone and 36,000 \rollouttwo trajectories for \glmdsxone.
\end{itemize}

Combined, our datasets contain over 200,000 trajectories, resulting in the largest open-source dataset for coding agents to date. We separate these trajectories by teacher model and rollout stage. For {\color{trajred}$\mathbf{T_2}$} trajectories, we further group them by verification threshold, with boundaries at $r = 0$, $0.25$, $0.50$, $0.75$, and $1$. We independently scale both {\color{trajred}$\mathbf{T_1}$} trajectories and {\color{trajred}$\mathbf{T_2}$} trajectories until a truncation ratio of $0.88$ is reached. Because there are multiple verification thresholds for {\color{trajred}$\mathbf{T_2}$} rollouts, we choose to scale $r = 0$ (completely unverified) trajectories, which has the highest data count. Our decision was influenced by experiments in Section \ref{sec:verification}, which indicate that completely unverified {\color{trajred}$\mathbf{T_2}$} rollouts are of equal or better quality than any verified rollouts.

Using \glmdsxone, we train SERA-32B and set a new state-of-the-art on SWE-Bench Verified for fully open-source 32B models evaluated at 32K context, with open-weight models like Devstral-Small-2-24B and larger models such as GLM-4.5-Air well within uncertainty bounds of one standard deviation. Evaluating at 64K context, SERA-32B again sets a state-of-the-art among fully open-source models, matching open-weight models such as FrogBoss-32B and only outperformed by Devstral-Small-2-24B \citep{rastogi2025devstral} among models with similar parameter counts. It is important to note that unlike these models, SERA-32B was not trained past 32K tokens and did not use any reinforcement learning, two factors that place it at a disadvantage at longer contexts. Still, SERA-32B performs extremely well and does not appear to have saturated yet.

\begin{figure}[!htb]
\centering
\includegraphics[width=\textwidth]{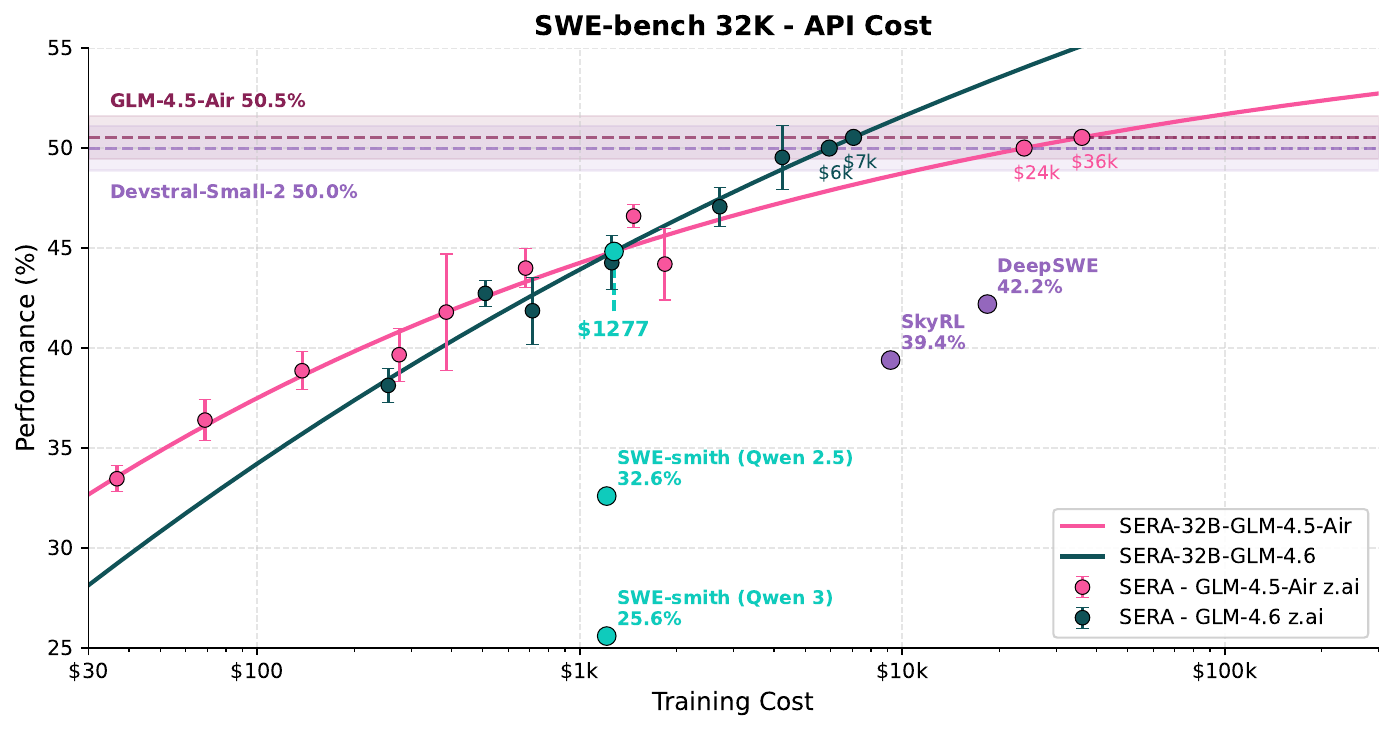}
\vspace{0.5em}
\includegraphics[width=\textwidth]{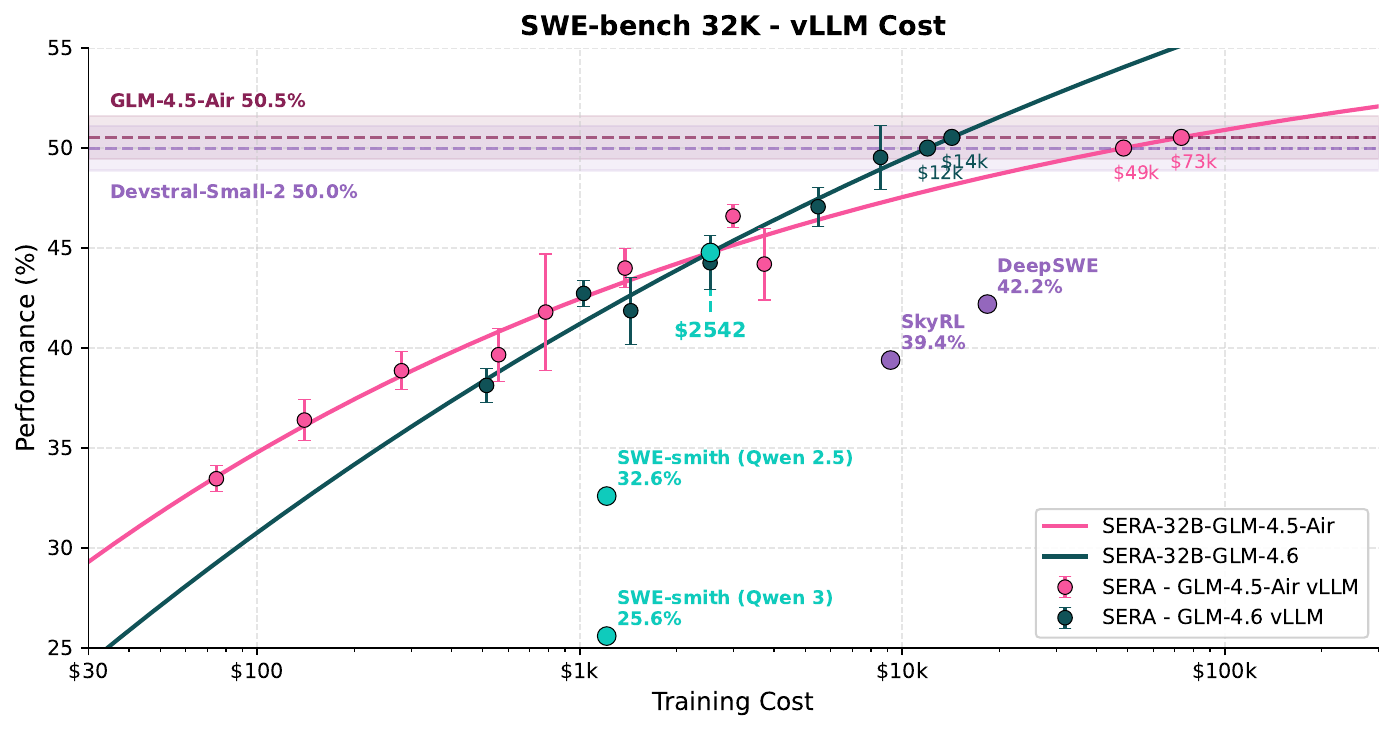}
\caption{Scaling and cost comparison of coding agent training approaches. \textbf{Top:} API cost when using z.ai with cached input pricing. \textbf{Bottom:} vLLM cost when self-hosting the teacher model. Horizontal lines indicate the cost at which our scaling law predicts matching Devstral-Small-2 and GLM-4.5-Air performance. Exact data points are provided in Table~\ref{tab:scaling_data}.}
\label{fig:scaling}
\end{figure}

We also train SERA-32B-GA using \glmairdsxone. While SERA-32B-GA lags behind SERA-32B, it still outperforms all other fully open-source models at 32K and 64K context lengths. Interestingly, SERA-32B-GA is able to match SERA-32B at low and intermediate sample sizes, after which point SERA-32B-GA's performance saturates. This suggests that the benefits of strong teacher models primarily emerge in high compute regimes. For researchers and practitioners, this means that it may be optimal to use a weaker teacher model depending on final performance goals and overall budget. Figure \ref{fig:scaling} highlights this crossover point, where the scaling curves for GLM-4.5-Air and GLM-4.6 intersect.

\subsection{Repository Specialization}
\label{sec:main-specialize}

\sera is the first synthetic data generation strategy that operates totally independent of a repository's unit tests. This allows users to rapidly specialize a base model to any downstream codebase, including private repositories. To emulate this scenario, we use \sera to generate data from the three largest repositories in SWE-Bench Verified: Django, Sympy, and Sphinx. Crucially, our synthetic training data is generated entirely independently of the evaluation instances, containing no information about the actual GitHub issues or their solutions in SWE-bench Verified. These repositories represent $231$ (46.2\%), $75$ (15.0\%), and $44$ (8.8\%) of the 500 instances in SWE-Bench Verified, respectively.

\textbf{Initial Data Generation:} Because every SWE-Bench instance is sourced from a unique commit, the set of instances from each repository will span multiple years. To account for this, we identify the earliest commit and latest commit in SWE-Bench Verified from each repository and generate data from five equally spaced commits in that period. While some functions are repeated across commits, each commit presents the codebase in a different context, which ensures trajectories in both the first and second rollout remain unique. Aggregating across commits, we obtain between 46,000 and 54,000 trajectories for each repository combined across both rollouts. We decide to train on both rollouts to increase sample size since the majority of the generated trajectories exceed 32K tokens. We investigate the effects of mixing rollouts in Section \ref{sec:mixing}. Due to compute constraints, we train on 8,000 trajectories per repository rather than the full dataset; however, we release all generated trajectories to enable future research to explore larger-scale specialization.

\textbf{Data Verification and Filtering:} We soft-verify {\color{trajred}$\mathbf{T_2}$} rollouts with a verification threshold of $0.5$. We cap {\color{trajred}$\mathbf{T_1}$} rollouts based on patch size and observations length. This selects against {\color{trajred}$\mathbf{T_1}$} rollouts that over-edit or make excessively long tool calls, a tendency that can quickly use up context. We find that this filtering significantly improves specialization performance, which we further investigate in Section \ref{sec:specialize}.

Finally, for each repository, we train on 3,000 soft-verified {\color{trajred}$\mathbf{T_2}$} rollouts and 5,000 filtered {\color{trajred}$\mathbf{T_1}$} rollouts. We note that these specific proportions were chosen based on preliminary experiments within our compute budget; a more systematic exploration of the optimal mixture would be valuable future work. In this setup, we match or exceed the teacher model GLM-4.5-Air on Django and Sympy instances, and also outperform Devstral-Small-2-24B (SoTA $\leq$32B parameters), while nearly matching their performance on Sphinx (Table \ref{tab:specialization}). This result is intuitive: the student encodes repository-specific knowledge in its weights, while the teacher can only access the codebase through its context window. We note that at 64K evaluation context, \sera underperforms baselines like Devstral-Small-2 because we train only at 32K context while these models are trained at 64K or longer; see Appendix~\ref{sec:specialization-64k} for 64K results. These results highlight that given the right data, it is possible to produce and even exceed state-of-the-art performance on specifically targeted repositories.

\begin{table}[!htb]
\centering
\begin{tabular}{l|ccc}
\toprule
    {\textbf{\small{Model}}} &
    {\textbf{\small{Django (231)}}} &
    {\textbf{\small{Sympy (75)}}} &
    {\textbf{\small{Sphinx (44)}}}
\\
\midrule
\rowcolor{pink!20}{SERA-32B-Django} & \textbf{52.23\% $\pm$ 1.64\%} & - & - \\
\rowcolor{pink!20}{SERA-32B-Sympy} & - & \textbf{51.11\% $\pm$ 1.54\%} & - \\
\rowcolor{pink!20}{SERA-32B-Sphinx} & - & - & 37.14\% $\pm$ 6.95\% \\
\midrule
\rowcolor{gray!20}{GLM-4.5-Air} & 51.20\% $\pm$ 1.80\% & 48.89\% $\pm$ 3.08\% & \textbf{43.51\% $\pm$ 0.58\%} \\
\rowcolor{gray!20}{Devstral-Small-2-24B} & 51.30\% $\pm$ 1.72\% & 47.56\% $\pm$ 4.68\% & 38.95\% $\pm$ 4.24\% \\
\bottomrule
\end{tabular}
\caption{Specialization results at 32K context comparing GLM-4.5-Air (teacher) and fine-tuned Qwen 3-32B (student) on the three largest repositories in SWE-Bench Verified. Fine-tuned models are trained on 8,000 synthetic trajectories from each repository. Results averaged over three seeds. Devstral-Small-2-24B results from \citet{rastogi2025devstral}. See Table~\ref{tab:specialization-64k} in the appendix for 64K evaluation results.}
\label{tab:specialization}
\end{table}

\begin{figure}[!htb]
\centering
\includegraphics[width=0.8\textwidth]{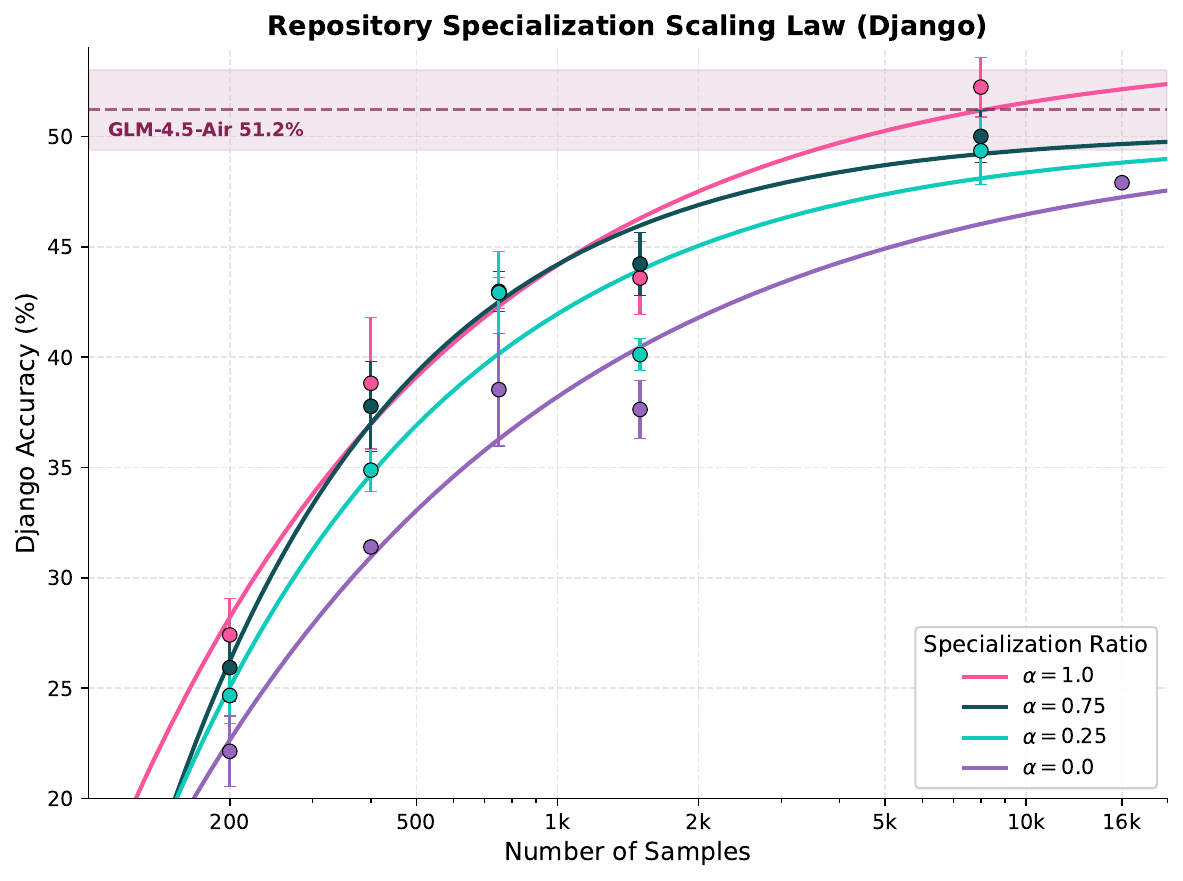}
\caption{Scaling law for repository specialization on Django. The specialization ratio $\alpha$ denotes the fraction of Django-specific data in the training mixture, with the remainder being general coding data. Dashed horizontal lines show the performance of GLM-4.5-Air and Devstral-Small-2 on Django instances, with shaded regions indicating $\pm 1$ standard deviation. With full specialization ($\alpha = 1.0$), the student model matches teacher performance at approximately 8,000 samples, significantly outperforming training on general data alone ($\alpha = 0.0$). Specialization performance increases with the ratio of Django-specific data.}
\label{fig:specialization}
\end{figure}

\textbf{Specialization Scaling Law:} To understand how data composition affects specialization efficiency, we fit scaling laws across different mixtures of Django-specific and general coding data (Figure~\ref{fig:specialization}). We define the specialization ratio $\alpha$ as the fraction of repository-specific data in the training mixture. At $\alpha = 1.0$ (pure Django data), the model matches teacher performance (GLM-4.5-Air at 51.2\%) with only 8,000 samples. In contrast, $\alpha = 0.0$ (pure general data) is unable to reach equivalent performance even at 16,000 samples. Intermediate mixtures ($\alpha = 0.75$, $\alpha = 0.25$) show increasing asymptotic performance as the proportion of specialized data increases. This indicates that when training for a target codebase, the ratio of specialized data is the most important factor. A one-way ANOVA reveals a statistically significant effect of specialization ratio at 1,500 samples, $F(3, 8) = 10.78$, $p = .003$, $\eta^2 = .80$. Post-hoc comparisons using Tukey's HSD showed significant differences between $\alpha = 1.0$ vs. $\alpha = 0.0$ ($p = .009$) and $\alpha = 0.75$ vs. $\alpha = 0.0$ ($p = .005$). This confirms that repository-specific data yields a statistically significant improvement over general data at equivalent sample sizes.


\section{Ablations and Analysis}
\label{sec:ablations}

In this section, we conduct comprehensive data ablations studying design choices in \sera. We focus on the impacts of verification; truncation; specialization; filtering; dataset mixing; and evaluation uncertainty. For these experiments, we draw data from \glmairdsxone.

\subsection{Verification}
\label{sec:verification}
\begin{figure}[!htb]
\centering
\includegraphics[width=0.8\textwidth]{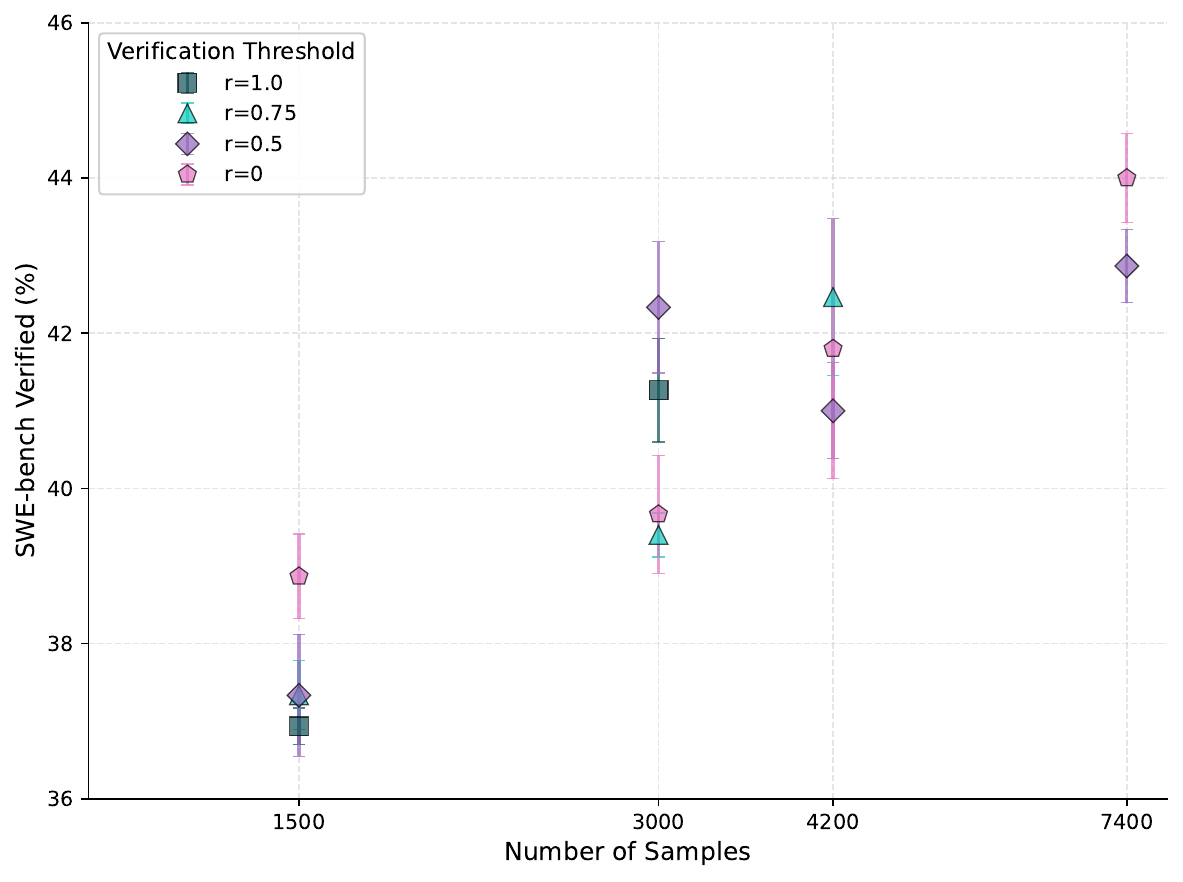}
\caption{Verification analysis comparing soft and hard verification approaches. Scaling curves show SWE-bench Verified performance for different verification thresholds ($r = 0.0, 0.25, 0.75, 1.0$) on \rollouttwo trajectories alongside unverified \rolloutone trajectories. All thresholds achieve similar performance at each scale, indicating that strict verification provides no significant benefit over soft or even unverified data.}
\label{fig:verification}
\end{figure}

In Figure \ref{fig:verification}, we ablate four different verification thresholds for \rollouttwo trajectories: $r = 0.0$, $0.25$, $0.75$, and $1$. We also plot the performance of models trained on \rolloutone trajectories (which are inherently unverified) at each scale for comparison. We study each verification threshold using only complete trajectories that fit within 32K tokens, with the final datapoint of each curve representing the largest possible dataset for that threshold. Only using trajectories $\leq$32K tokens reduces the total number of trainable trajectories but allows us to avoid confounding factors introduced by truncation. 

If verification was essential for performance, we would expect to observe increasing performance as the verification threshold increases. Instead, scaling up to 7,400 samples, we find that all verification thresholds perform similarly. For example, at the maximum scale, training on \rollouttwo trajectories that are soft-verified at $r= 0.5$ shows no benefit over training on completely unverified \rollouttwo trajectories. Furthermore, models trained on \rolloutone trajectories from the first rollout result in similar SWE-Bench Verified performance at each scale, despite representing a completely different distribution of coding tasks. 

These results suggest that verification is not a necessity for high quality coding data, a behavior that is similar to what has been observed in other types of reasoning tasks \citep{chandra2026shapethoughtdistributionmatters}. We hypothesize that this is because even incorrect trajectories are can contain important skills, such as how to convert an intention into a relevant code edit, even if the intention does not perfectly address the PR at hand. Indeed, a Kruskal-Wallis H-test reveals no statistically significant difference between verification thresholds, $H(3) = 7.19$, $p = .066$, $\varepsilon^2 = .52$. This indicates that soft verification performs as well as hard verification, and even unverified data achieves comparable results.

\subsection{Line-Level Recall}
\label{sec:llr}

\begin{figure}[!htb]
\centering
\includegraphics[width=0.55\textwidth]{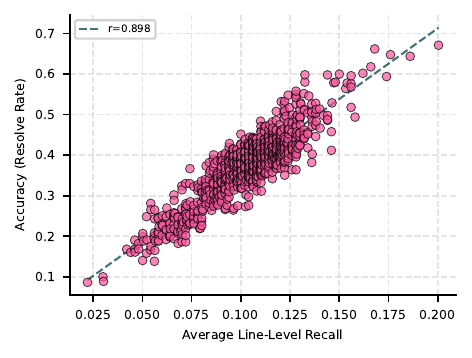}
\caption{Scatterplot comparison of issue resolution rates and linelevel recall across 1,036 SWE-Bench Verified evaluations using a mixture of open and closed weight models.}
\label{fig:llr}
\end{figure}

Verification and reinforcement learning may become necessary once a model saturates on skills that unverified trajectories teach. In this data regime, an important question
is whether line-level recall can be used to measure code
correctness. We find that the answer is a resounding yes. Across 1,036 SWE-Bench Verified evaluations, comprised of over 500,000 individual tasks, we find an extremely high correlation between line-level recall and unit test resolution, illustrated in Figure \ref{fig:llr}. This suggests that line-level recall can serve as a continuous proxy for code correctness, while having the added benefits of being cheaper, quicker to run, and generalizable to any codebase.

\subsection{Truncation}
\label{sec:trunc}

Truncation is a popular but understudied practice across coding agent research to handle trajectories from teacher models whose context windows are larger than the base model's. Without truncation, base model context limits make it impossible to train on significant amounts of data: $23.26\%$ of SWE-smith and $24.83\%$ of \glmairdsxone (verified at $r = 1$) exceed 32K tokens.

As a result, current SFT methods will slice long trajectories to fit inside the base model's context window. While this allows every data sample to be used, it assumes that all sliced trajectories are similar quality. We hypothesize that this is a faulty assumption. For example, a trajectory that represents only $50\%$ of a trajectory's total steps is intuitively more noisy than a sliced trajectory that represents $95\%$ of a trajectory's total steps.

To test this hypothesis, we order \rolloutone trajectories from \glmairdsxone based on the ratio of trajectory steps that fit in 32K tokens, a property we term ``truncation ratio''. We partition the ordered \rolloutone trajectories into subsets of 3,000 samples each. This forces each subsequent partition to contain trajectories with strictly lower truncation ratios than the previous partition. We then train Qwen3-32B on each partition. \rolloutone trajectories work well because they are longer than \rollouttwo trajectories on average while exhibiting similar scaling trends. This allows us to study the effect of training on a wide range of truncation ratios with a non-trivial amount of data and expect findings to translate.

\begin{figure}[!htb]
\centering
\includegraphics[width=0.7\textwidth]{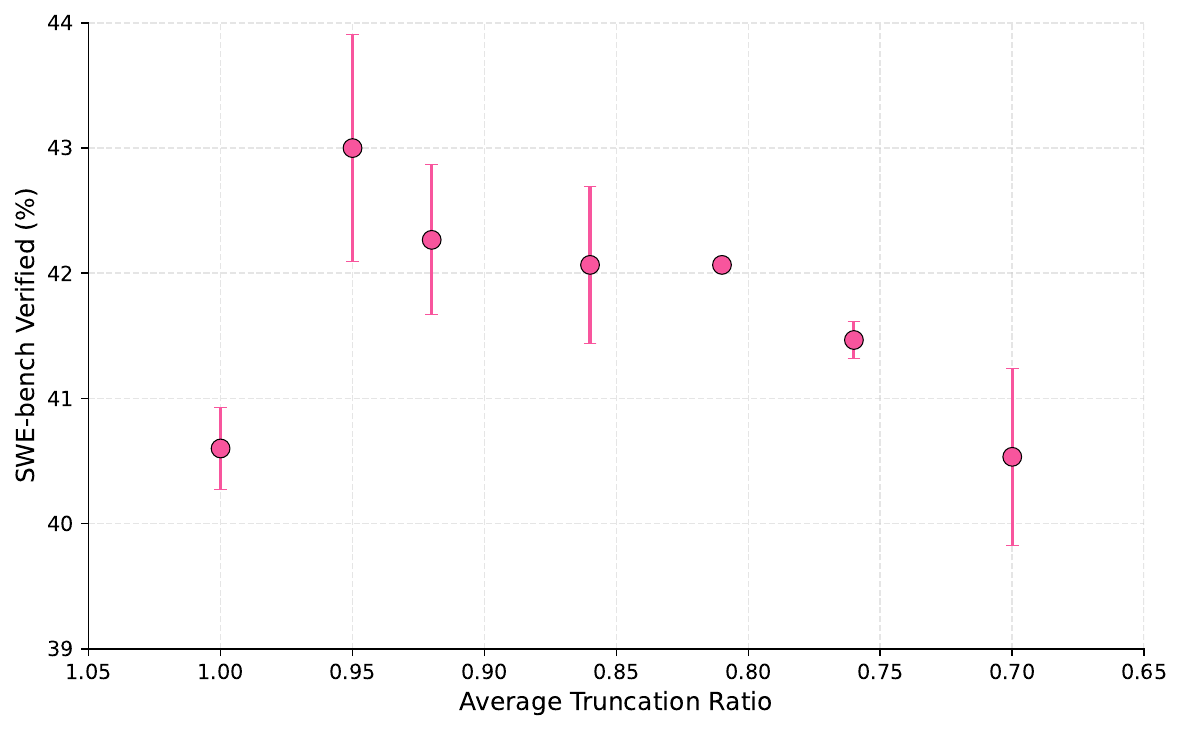}
\caption{SWE-bench Verified performance vs. truncation ratio at 32K context. Each point represents 3,000 \rolloutone trajectories partitioned by truncation ratio, averaged over 3 seeds. Trajectories with truncation ratio 0.95 perform best.}
\label{fig:truncation-curve}
\end{figure}

In Figure \ref{fig:truncation-curve} we plot SWE-Bench Verified performance against the average truncation ratio from every partition. Surprisingly, we find that the best data comes from trajectories that have high truncation ratios but are not fully contained in 32K tokens. Subsequent truncation ratios result in gradually decreasing performance. We suspect that this is due to a combination of factors, such as longer trajectories reflecting more difficult tasks and that a model's final steps are typically focused on the redundant task of submitting its solution instead of problem solving.

We further explore this phenomenon in Table \ref{tab:truncation-study}, where we compare the effect of selecting for high truncation ratios against the current practice of using sliced trajectories arbitrarily. We find that randomly picking sliced trajectories results in much lower performance than curating for high truncation ratios. This suggests that existing truncation approaches are suboptimal and may hurt performance. These results inform our scaling experiments in Section \ref{sec:scaling}, where we extend our datasets by first ordering by truncation ratio and then truncating until a ratio of $0.88$ is reached. We conservatively pick this threshold to avoid deterioration and leave further investigation for future work.
\begin{table}[h!]
\centering
\begin{tabular}{c|c|c}
\toprule
\textbf{Fully Fit in 32K} & \textbf{Ordered Truncated} & \textbf{Random Truncated} \\
\midrule
40.60\% $\pm$ 0.69\% & \textbf{43.00\% $\pm$ 1.93\%} & 37.47\% $\pm$ 0.50\% \\
\bottomrule
\end{tabular}
\caption{Effect of truncation strategy on SWE-bench Verified performance. All conditions use 3,000 \rolloutone trajectories from GLM-4.5-Air, trained on Qwen 3-32B and evaluated at 32K context. Ordered truncation selects trajectories with the highest truncation ratios $< 1$, while random truncation samples arbitrarily from all trajectories exceeding 32K tokens. Results averaged over 3 seeds.}
\label{tab:truncation-study}
\end{table}

Assumptions were met for an independent samples t-test, which showed that ordered truncation ($M = 43.00$, $SD = 1.93$) significantly outperforms random truncation ($M = 37.47$, $SD = 0.50$), $t(4) = 4.81$, $p = .009$, $d = 3.93$. This confirms that the order in which content is truncated matters: preserving earlier turns in the trajectory is more effective than random truncation.

\subsection{Data Filtering for Specialization}
\label{sec:specialize}

\textbf{Long Edit and Tool Call Length Filtering:} Table \ref{tab:filtering} highlights the effect of filtering out trajectories with long edits and large tool calls during specialization. We classify a excessively long edit as final edits exceeding $40$ lines and excessively large tool calls as tool call responses containing more than $600$ tokens. Ablating these filtering conditions for each repository, we find that no single filter setting reliably generalizes for all repositories. Indeed, filtering long edits works very well for Django and Sympy, but is ineffective for Sphinx, which instead benefits from filtering for tool call size.
\begin{table}[h!]
\centering
\begin{tabular}{l|c|c|c}
\toprule
\textbf{Repository} & \textbf{No Filter} & \textbf{Patch $\leq$40 Lines} & \textbf{Both Filters} \\
\midrule
Django & 49.93\% $\pm$ 1.64\% & \textbf{52.23\% $\pm$ 1.64\%} & 50.07\% $\pm$ 2.95\% \\
Sympy & 46.67\% $\pm$ 2.67\% & \textbf{51.11\% $\pm$ 1.54\%} & 44.89\% $\pm$ 1.54\% \\
Sphinx & 32.29\% $\pm$ 1.94\% & 30.30\% $\pm$ 6.95\% & \textbf{37.14\% $\pm$ 6.95\%} \\
\bottomrule
\end{tabular}
\caption{Effect of filtering on repository-specialized data. ``Patch $\leq$40 Lines'' drops trajectories with patches exceeding 40 lines. ``Both Filters'' additionally removes trajectories where average tool output exceeds 600 tokens. Filtering patches improves Django (+2.3\%) and Sympy (+4.4\%), while Sphinx benefits from the combined filter (+4.9\%). Trained on Qwen 3-32B, evaluated at 32K context. Results averaged over 3 seeds.}
\label{tab:filtering}
\end{table}

We also apply these filtering techniques to \rolloutone trajectories from \glmairdsxone, as shown in Table \ref{tab:filtering-general}, which results in no significant improvements.

\begin{table}[h!]
\centering
\begin{tabular}{c|c|c}
\toprule
\textbf{No Filter} & \textbf{Patch $\leq$40 Lines} & \textbf{Tool Output $\leq$600} \\
\midrule
43.93\% $\pm$ 1.30\% & 43.67\% $\pm$ 1.60\% & 44.00\% $\pm$ 0.00\% \\
\bottomrule
\end{tabular}
\caption{Effect of filtering on general \rolloutone trajectories from GLM-4.5-Air. ``Patch $\leq$40 Lines'' removes trajectories with patches exceeding 40 lines (n=5,364 from n=7,400). ``Tool Output $\leq$600'' removes trajectories where average tool output exceeds 600 tokens (n=6,136 from n=7,400). Neither filter improves performance on general data. Trained on Qwen 3-32B, evaluated at 32K context. Results averaged over 3 seeds.}
\label{tab:filtering-general}
\end{table}

Taken together, our results suggest that filtering can be targeted to improve performance on specific repositories but is not as reliable in aggregate. The effectiveness of filtering methods likely reflects individual codebase characteristics. As a result, we suggest that users develop their own filtering heuristics for personal repositories.

\textbf{Specializing to Multiple Repositories:} We also jointly train on Django and Sympy to investigate whether SERA can be used to specialize to multiple codebases at once. We randomly sample half our initial Django and Sympy datasets from Section \ref{sec:main-specialize}, training on the combined dataset. We evaluate Django and Sympy instances separately, then average them with equal weighting, which debiases against the larger number of Django instances in SWE-Bench Verified. We find that while performance drops slightly on each constituent codebase, the average performance of the combined dataset outperforms 10,000 \rolloutone trajectories from \glmairdsxone (Table~\ref{tab:multi-repo}). In addition, independent performances on Django and Sympy still compare favorably to the teacher model. This indicates that \sera can be applied to multiple codebases for broadly improved performance, which reflects the needs of enterprises and larger research teams.
\begin{table}[h!]
\centering
\begin{tabular}{l|c|c|c}
\toprule
\textbf{Training Data} & \textbf{Django} & \textbf{Sympy} & \textbf{Average} \\
\midrule
Specialized (8k Django) & \textbf{52.23\% $\pm$ 1.64\%} & --- & --- \\
Specialized (8k Sympy) & --- & \textbf{51.11\% $\pm$ 1.54\%} & --- \\
Specialized (4k Django + 4k Sympy) & 50.07\% $\pm$ 0.50\% & 46.22\% $\pm$ 3.35\% & \textbf{48.15\%} \\
General (10k samples) & 45.60\% $\pm$ 0.90\% & 48.89\% $\pm$ 0.77\% & 47.25\% \\
\bottomrule
\end{tabular}
\caption{Multi-repository specialization results. Specialized training on 8,000 single-repository trajectories achieves the best per-repository performance. Mixed training (4,000 Django + 4,000 Sympy) achieves the best average. General training uses 10,000 \rolloutone trajectories from \glmairdsxone. Average is computed with equal weighting between Django and Sympy. Trained on Qwen 3-32B, evaluated at 32K context. Results averaged over 3 seeds.}
\label{tab:multi-repo}
\end{table}

\subsection{Teacher Models}
\label{sec:teachers}

 In Appendix~\ref{sec:additional_baselines}, we show that GLM-4.5-Air is a much better teacher than Claude 3.7 Sonnet despite similar SWE-Bench Verified performance. We hypothesize that this is in part due to GLM-4.5-Air's reasoning traces, which are longer and significantly more elaborate.

To study the effect of reasoning traces, we train on 4,200 \rollouttwo trajectories from GLM-4.5-Air where we remove reasoning traces and leave only tool calls. In Table \ref{tab:reasoning_length} we find that this significantly degrades performance compared to the unchanged trajectories. Our results confirm our hypothesis that high-quality reasoning traces are essential when distilling data for coding agents.

\begin{table}[h!]
\centering
\begin{tabular}{l|c}
\toprule
\textbf{Condition} & \textbf{SWE-bench Verified} \\
\midrule
With Reasoning & \textbf{41.00\% $\pm$ 1.31\%} \\
No Reasoning & 23.00\% \\
\bottomrule
\end{tabular}
\caption{Effect of reasoning traces on coding agent performance. Both conditions use 4,200 \rollouttwo trajectories from GLM-4.5-Air. No Reasoning removes all reasoning traces, retaining only tool calls. Trained on Qwen 3-32B, evaluated at 32K context. With Reasoning results averaged over 3 seeds.}
\label{tab:reasoning_length}
\end{table}

\subsection{Rollout Mixing}
\label{sec:mixing}

In Section \ref{sec:main-specialize}, we mix \rolloutone and \rollouttwo trajectories to increase sample size during specialization. We repeat this experiment at a significantly larger scale in Table \ref{tab:mixing} using GLM-4.6 as a teacher. Combining 16,000 \rollouttwo trajectories and 9,224 \rolloutone trajectories improves performance compared to training only on 16,000 \rollouttwo trajectories. While the resulting model falls just shy of scaling only \rollouttwo trajectories, the results suggest that \rolloutone and \rollouttwo can be reliably mixed to extract further performance gains in data constrained settings.

\begin{table}[h!]
\centering
\begin{tabular}{l|c|c|c}
\toprule
& \textbf{16k \rollouttwo} & \textbf{25k \rollouttwo} & \textbf{16k \rollouttwo + 9k \rolloutone} \\
\midrule
SWE-bench Verified & 47.07\% $\pm$ 1.21\% & \textbf{49.53\% $\pm$ 1.94\%} & 49.00\% $\pm$ 1.64\% \\
\bottomrule
\end{tabular}
\caption{Effect of mixing \rolloutone and \rollouttwo trajectories. All data generated using GLM-4.6 as teacher. 16k \rollouttwo uses 16,000 second-rollout trajectories. 25k \rollouttwo uses 25,224 second-rollout trajectories. Mixed condition combines 16,000 \rollouttwo with 9,224 \rolloutone trajectories. Mixing improves over 16k \rollouttwo alone (+1.9\%) but falls slightly short of scaling to 25k \rollouttwo. Trained on Qwen 3-32B, evaluated at 32K context. Results averaged over 3 seeds.}
\label{tab:mixing}
\end{table}

We run a one-way ANOVA test revealing no statistically significant difference when trajectories are mixed compared to when they are not: $F(2, 6) = 1.91$, $p = .229$, $\eta^2 = .39$. This further indicates that \rolloutone and \rollouttwo trajectories can be combined with little to no performance degradation, enabling improved sampling efficiency.

\section{Robustness of Evaluations}
\label{sec:eval_distribution}

To assess the reliability of our findings, we conducted a systematic statistical analysis across all experiments in this paper. Our analysis aggregates within and between experiments and for multiple random seeds that include all experiments for scaling laws, verification thresholds, truncation strategies, specialization mixtures, filtering ablations, and baseline comparisons. In total, this analysis covers 78 experimental conditions, each evaluated with three random seeds, yielding 234 individual evaluation runs. Based on our findings we concluded with recommended best practices for coding agent evaluations.

\textbf{Observed Variance:} Across all experimental conditions, we observe standard deviations ranging from 0.5\% to 3.0\%, with a median of 1.2\%. This is problematic when the magnitude of improvement in coding agent research is typically also 1--3\%. Many reported gains in the literature fall within one standard deviation of run-to-run noise.

\textbf{Signal-to-Noise Analysis:} A practical way to assess the validity of observed improvements is to compute the signal-to-noise ratio (SNR): the absolute difference between methods divided by the typical run-to-run variance. When SNR $< 1$, noise dominates and the result cannot be trusted. When SNR is between 1--2, the result is borderline and requires more seeds. When SNR $> 2$, there is likely a significant effect. Applying this framework to our experiments:
\begin{itemize}
    \item \textbf{High confidence (SNR $> 3$):} Specialized vs.\ general data (+4.3\%, SNR=5.6), SERA vs.\ SWE-smith with same teacher (+4.7\%, SNR=4.4), scaling law predictions (mean error 0.4\%)
    \item \textbf{Moderate confidence (SNR 2--3):} Verification threshold equivalence (all within 2.9\%, SNR confirms no difference), truncation ratio effects (+2.4\%, SNR=2.2)
    \item \textbf{Low confidence (SNR $< 2$):} Student matching teacher at 8k samples (1.7\% difference, SNR=1.4, error bars overlap)
\end{itemize}

\textbf{How Many Seeds Do You Need?} Based on the empirical variance in our data (median standard deviation of 1.2\%), Table~\ref{tab:seed_requirements} shows approximately how many seeds are required to achieve SNR $\geq 2$ for different effect sizes. These estimates follow directly from the definition: to achieve SNR $= 2$ for an effect of size $\delta$, the standard error must be at most $\delta/2$, requiring $n \approx (2 \cdot \text{std} / \delta)^2$ seeds.

\begin{table}[h!]
\centering
\begin{tabular}{c|c|l}
\toprule
\textbf{Effect Size} & \textbf{Seeds for SNR $\geq 2$} & \textbf{Reliability with $n=3$} \\
\midrule
1\% & $\sim$15 & Cannot detect reliably \\
2\% & $\sim$4 & Borderline \\
3\% & $\sim$2 & Adequate \\
5\% & $\sim$2 & High confidence \\
\bottomrule
\end{tabular}
\caption{Seeds required to achieve SNR $\geq 2$ for different effect sizes, derived from the empirical variance in our experiments (median std = 1.2\%). With only 3 seeds, improvements below 2--3\% should be treated with skepticism.}
\label{tab:seed_requirements}
\end{table}

\textbf{The Single-Seed Problem:} Many published results in coding agent research report single-seed evaluations. Our data demonstrates the danger of this practice. Across multiple experiments, we find cases where different random seeds lead to opposite conclusions about which method is best. For example, in our truncation experiments, seeds 1 and 2 identify ratio 0.95 as optimal, while seed 3 identifies ratio 0.92 as optimal with 0.95 performing 2.2\% worse. Single-seed ablations cannot be trusted.

\textbf{Cross-Model Generalization:} An concerning observation is that methods might not generalize well across different base models or teacher models. We observe that SWE-smith achieves 32.6\% with Qwen 2.5-32B but only 25.3\% with Qwen 3-32B. This 7.3\% drop suggesting the method may have been unintentionally optimized for the earlier model family. For our method, changing the teacher model to Sonnet 3.7 and Sonnet 4.0 behaves as expected, demonstrating cross-model generalization. However, we did not have resources to test cross-model generalizations for the base model. Even our findings should be interpreted with caution: improvements we observe may not transfer to base models outside the Qwen and GLM families.

\textbf{Scaling Laws as a Robustness Check:} We found scaling laws to be invaluable for ensuring reliable results and recommend that future work in coding agents incorporate them where possible. Our scaling experiments (Figure~\ref{fig:scaling}) show that performance follows a highly predictable power law ($R^2 > 0.95$, mean prediction error 0.4\%). Scaling laws provide several benefits: (1) \emph{experimentation efficiency}: running experiments at smaller, cheaper scales while extrapolating findings to larger scales, since power laws have proven predictable and reliable; (2) \emph{cost estimation}: predicting the resources required to reach target performance levels before committing to expensive runs; (3) \emph{method comparison}: estimating sample efficiency and cost differences between methods without running exhaustive experiments at all scales; and (4) \emph{robustness checking}: when a method's performance falls significantly outside the scaling law prediction, this signals either a genuine breakthrough or, more likely, noise or overfitting to a particular configuration.

\textbf{Recommendations:} Based on this analysis, we encourage researchers to (1) run a minimum of 3 seeds, preferably more for ablations expecting improvements below 3\%, (2) report standard deviations alongside means, (3) compute the signal-to-noise ratio and treat SNR $< 2$ results as preliminary, (4) verify that improvements transfer across model configurations, and (5) fit scaling laws where feasible to enable efficient experimentation and robustness checking.

\section{Deployment}
\label{sec:deployment}

As part of this release, we provide a lightweight proxy server that enables Claude Code to use \sera as its backend. This section describes implementation considerations for deploying \sera in practice.

\textbf{Tool Format Compatibility:}
\sera is trained on SWE-agent tool formats and performs best with exact format matching at inference time. Deploying the model with a different agent scaffold, or even subtle formatting differences, degrades performance significantly. Claude Code uses a different tool set (\texttt{Read}, \texttt{Edit}, \texttt{Write}, \texttt{Bash}) than SWE-agent (\texttt{str\_replace\_editor}, \texttt{bash}), so our sera-cli proxy translates between them. Path normalization is also required: data generation with SWE-agent uses a consistent working directory across all trajectories, so the proxy translates these paths to the user's current working directory. Tool result formatting must match training exactly including details like whitespace conventions and directory listing formats. When any of these conventions mismatch, the model can enter unproductive loops (e.g., repeatedly verifying edits that were already applied correctly), resulting in a poor experience for users. These issues are difficult to detect without agent scaffold specific evaluations.

\textbf{Infrastructure:}
The proxy connects Claude Code to any OpenAI-compatible endpoint serving \sera. We use vLLM~\citep{kwon2023vllm} with the Hermes tool calling parser. For serverless deployment, we provide Modal integration scripts, though the model is portable to any cloud GPU provider or on-premises infrastructure. The proxy handles API translation (Anthropic format to OpenAI format), tool mapping, and response streaming and can be easily modified to handle \sera models specialized on a specific repository. SERA-32B requires at least one 80GB GPU (e.g., A100 80GB, H100, or greater) for deployment. Quantization (e.g., AWQ, GPTQ) can further improve throughput and reduce memory requirements.

\section{Related Work}

Training data generation has emerged as a critical bottleneck for developing capable software engineering agents. Several approaches address this challenge through different methodologies for environment construction, data synthesis, and verification.

The most closely related work are other synthetic data generation approaches. such work includes, BugPilot~\citep{sonwane2025bugpilot} which synthesizes bugs by instructing agents to add features, capturing unintentional test breakages as training data that more closely mirrors real development patterns. SWE-Synth~\citep{pham2025swesynth} leverages LLM agents to simulate debugging workflows, producing bug-fix pairs with test cases and structured repair trajectories. SWE-Mirror~\citep{wang2025swemirror} takes a different approach, distilling real issues from GitHub and mirroring them into repositories with configured environments, enabling data generation across multiple programming languages. SWE-smith~\citep{yang2025swesmith} introduces an automated pipeline that synthesizes task instances by breaking existing tests in Python codebases, producing 50K instances from 128 repositories. R2E-Gym~\citep{jain2025r2egym} introduces SYNGEN, a synthetic data curation recipe that uses test generation and commit back-translation to scale environment curation without relying on human-written issues. Skywork-SWE~\citep{zeng2025skyworkswe} investigates data scaling laws for software engineering, demonstrating that model performance continues to improve with dataset size without saturation.

Environment construction has been addressed through multiple strategies: SWE-Gym~\citep{pan2024swegym} provides the first dedicated training environment with over 2,400 executable Python task instances for training language model-based software engineering agents. Repo2Run~\citep{hu2025repo2run} uses LLM agents to iteratively build Docker environments from repository feedback, SWE-Factory~\citep{guo2025swefactory} employs multi-agent collaboration with environment memory pools, and RepoST~\citep{xie2025repost} leverages sandbox testing to isolate functions and their dependencies for scalable construction.

Recent work has also explored optimizing the training process itself. SWE-Lego~\citep{tao2026swelego} demonstrates that a refined supervised fine-tuning procedure with error masking and difficulty-based curriculum can achieve state-of-the-art performance without reinforcement learning. SWE-Playground~\citep{zhu2025sweplayground} synthetically generates projects and tasks from scratch using strong language models, eliminating reliance on external data sources. Agent Data Protocol~\citep{song2025adp} introduces a unified representation language for agent training datasets, enabling standardized fine-tuning across heterogeneous data sources from coding, browsing, and tool-use domains.

\textbf{Evaluation Benchmarks:}
Standardized benchmarks have been essential for measuring progress in automated software engineering. SWE-bench~\citep{jimenez2024swebench} established the primary evaluation framework using real GitHub issues and pull requests from popular Python repositories. The benchmark has since been extended: SWE-rebench~\citep{badertdinov2025swerebench} addresses contamination through continuous collection of fresh tasks, and Multi-SWE-bench~\citep{zan2025multiswebench} expands coverage to multiple programming languages including Java, TypeScript, Go, and Rust. Beyond repository-level tasks, BigCodeBench~\citep{zhuo2024bigcodebench} evaluates models on tasks requiring diverse function calls from 139 libraries, while CodeRAG-Bench~\citep{wang2024coderag} systematically studies how retrieval-augmented generation can improve code generation across basic programming to repository-level problems.

\textbf{Agent Architectures:}
Various architectural approaches have been proposed for software engineering agents. SWE-agent~\citep{yang2024sweagent} introduces a custom agent-computer interface designed for efficient repository navigation and code editing. OpenHands~\citep{wang2024openhands} provides a modular platform supporting multiple agent implementations with standardized tool interfaces and sandboxed execution. Agentless~\citep{xia2024agentless} demonstrates that competitive performance can be achieved through a simpler three-phase approach combining localization, repair, and validation, without complex agent scaffolding. OpenHands-Versa~\citep{soni2025openhandsversa} further shows that a generalist agent with minimal tools---code editing, web search, and multimodal browsing---can achieve competitive performance across diverse benchmarks without domain-specific specialization.

\textbf{Training Methods:}
Beyond data generation, several works explore training approaches for software engineering models. SWE-RL~\citep{wei2025swerl} applies reinforcement learning on open software evolution data using lightweight rule-based rewards to improve reasoning capabilities. SkyRL-Agent~\citep{cao2025skyrl} provides efficient RL training infrastructure for multi-turn LLM agents. SWE-Fixer~\citep{xie2025swefixer} combines BM25-based file retrieval with a separate code editing module, training both components on 110K GitHub issues. These reinforcement learning approaches require substantial infrastructure for online rollouts and distributed training, which we discuss in Section~\ref{sec:background}.

\section{Limitations}

There are several limitations of our work. While we draw certain best-faith conclusions based on our empirical results, this section is mostly about uncertainty and what this might mean for the interpretation of our results. We also highlight the gaps that might make our conclusions one-sided or biased due to the particular experiments that we ran.

\textbf{Hard vs Soft Verification:} We show that different verification levels, including no verification at all, perform approximately equally well. Conceptually, this is surprising. One explanation is that early performance gains on coding tasks depend primarily on learning skills like converting intentions into code edits and navigating codebases, rather than on code correctness. However, once a model saturates on these aspects, verified correct code may become necessary for further improvement. We could not test this hypothesis at our scale. It is possible that with larger models or more training data, soft verification no longer suffices and hard verification with correct code becomes essential.

\textbf{Matching Teacher Performance:} Our specialization results show that we can match or exceed teacher performance at around 8,000 samples per repository, and our scaling laws predict this trend continues with sufficient data. However, we could not verify whether this advantage scales further due to compute limitations. The practical takeaway may be that while exceeding teacher performance is possible, the gains are modest and likely level off. At that point, upgrading to a stronger teacher becomes more efficient than generating additional data.

\textbf{Evaluation only on SWE-bench:} We evaluate only on SWE-bench Verified. When using our model for our own coding tasks, we find it performs well but exhibits some undesirable behaviors leftover from training. For instance, it attempts to call a nonexistent submit tool when its finished editing in Claude Code. While this suggests our results may generalize to some degree, we have not validated our model on other coding benchmarks or tasks, and we do not know how well it performs more broadly.

\textbf{Private Repository Specialization :} We demonstrate specialization on Django, Sympy, and Sphinx because these repositories have test data that allows us to evaluate whether specialization works. However, these are public repositories likely included in base model training data. Our specialization experiments may therefore be biased. While specialization effects are well-studied in fine-tuning scaling laws and our results appear plausible, we have not verified specialization on truly private codebases that models have never seen because we have no evaluation data to test this directly.

\textbf{Statistical Robustness:} As discussed in Section~\ref{sec:eval_distribution}, some of our comparisons are underpowered with $n=3$ seeds. Some reported effects may be noise rather than genuine improvements. We encourage readers to focus on large effects (>3\%) and treat smaller differences with appropriate skepticism.

\textbf{Model-Specific Results:} All experiments use Qwen-3 family of models as the base model and GLM-4.5-Air or GLM-4.6 as teachers. While we have some experiments with Claude 3.7 Sonnet and Claude 4.0 Sonnet that hint at generalization of our method, we do not know whether our findings generalize to other model families when evaluated thoroughly. The concerns we raise about model-specific optimization in the evaluation section apply equally to our own work.

\section{Broader Impact}

We believe that the release of the Ai2 Open Coding Agents that include our SERA models will have significant impact on the research community by enabling research on coing agents without requiring large resources or complicated systems. We also believe private specialization will have a significant effect of how small organizations use coding agents. In this section, we extend these discussion and the effects our work has.

\textbf{Democratizing Coding Agent Research:} A central barrier to progress in coding agent research has been the prohibitive cost and infrastructure complexity required to train competitive models. The Ai2 Open Coding Agents initiative aims to remove these barriers, with \sera as its first release. Reinforcement learning approaches require teams of 12 or more researchers, clusters of 64+ GPUs, and months of engineering effort to build sandboxed execution environments. Our work with \sera presents a much less resource-intensive approach. This shift makes coding agent research feasible for individual researchers, small academic labs, and institutions in regions without access to large-scale compute infrastructure. By releasing 200,000 trajectories, our training code, we aim to further lower the barrier to entry so that the study of coding agents is not concentrated among a handful of well-resourced industry labs.

\textbf{Enabling Private Codebase Specialization:} The ability to specialize a coding agent to a private codebase has significant implications for individuals and small companies. Currently, developers who want AI-assisted coding must send their proprietary code to cloud API providers, creating privacy and intellectual property concerns while these closed system are also not adapted to work well with private data. \sera enables a fundamentally different workflow: users can train a small, local model specialized to their own codebase without exposing their code to any third party. This is particularly relevant for startups with proprietary algorithms, regulated industries (healthcare, finance, defense) where code cannot leave secure environments, and open-source maintainers who want AI assistance tailored to their specific project conventions.

\textbf{Open Science and Reproducibility:} We release all components needed to reproduce and extend our work: training data, generation code, model weights, and evaluation scripts. Beyond enabling replication, this provides a shared foundation that other researchers can build upon without recreating expensive infrastructure from scratch. Our detailed cost analyses and scaling laws further serve the community by providing realistic expectations for resource planning and by identifying which experimental factors actually matter, potentially saving other groups from pursuing unproductive directions.

\subsubsection*{Acknowledgments}
This research is supported by Schmidt Sciences and by a Laude Institute Slingshot. We thank Taira Anderson, Caroline Wu, Johann Dahm, Sam Skjonsberg, David Albright, Kyle Wiggers, Hanna Hajishirzi, Ranjay Krishna, Crystal Nam, and the Beaker Team for their feedback and support.

\clearpage
\bibliographystyle{abbrvnat}

\clearpage
\bibliography{references}
\include{appendix}

\end{document}

%% file: appendix.tex
\appendix

\section{Scaling Law and Data Points}
\label{sec:scaling_data}

We fit a power law to our cost-performance data to predict how SERA scales with additional investment. The scaling law takes the form:
\begin{equation}
y = c - a \cdot x^{-b},
\label{eq:scaling_law}
\end{equation}
where $y$ is the SWE-bench Verified resolve rate (\%), $x$ is the total training cost in thousands of dollars (including both data generation and training), $c$ is the asymptotic performance ceiling as cost approaches infinity, $a$ is a scaling coefficient controlling how far below the asymptote performance begins, and $b$ is the power law exponent governing the rate of diminishing returns. The curve is fitted separately for each cost regime (vLLM self-hosting at \$0.187/sample and z.ai API at \$0.092/sample), yielding different $(c, a, b)$ parameters since the same number of samples maps to different costs.

To predict the cost of matching a baseline system, we solve Equation~\ref{eq:scaling_law} for $x$ at the target performance level $y^*$:
\begin{equation}
x^* = \left(\frac{a}{c - y^*}\right)^{1/b}.
\end{equation}
For example, Devstral-Small-2 achieves 50.0\% and GLM-4.5-Air achieves 50.5\% on SWE-bench Verified. Solving for these targets yields predicted costs of \$7K (z.ai API) or \$15K (vLLM) to match Devstral-Small-2, and \$9K (z.ai API) or \$19K (vLLM) to match GLM-4.5-Air. The fitted asymptote is approximately 70\%, suggesting substantial headroom remains if data quantity is scaled further, though we note this extrapolation is uncertain as it extends well beyond our observed data range.

Table~\ref{tab:scaling_data} provides the exact data points underlying the scaling law in Figure~\ref{fig:scaling}. All experiments use Qwen 3-32B as the base model trained on SERA data generated with GLM-4.5-Air as the teacher, evaluated on SWE-bench Verified at 32K context length. Each condition is evaluated over 3 random seeds. We report these values to enable other researchers to directly compare against our scaling curve without needing to read approximate values from the plot.

\begin{table}[h!]
\centering
\caption{Exact scaling law data points for Figure~\ref{fig:scaling}. Performance is SWE-bench Verified resolve rate (\%). Costs include both data generation and training. Per-sample cost is \$0.187 for vLLM self-hosting and \$0.092 for the z.ai API.}
\label{tab:scaling_data}
\begin{tabular}{r|ccc|c|c|r|r}
\toprule
\textbf{Samples} & \textbf{Seed 1} & \textbf{Seed 2} & \textbf{Seed 3} & \textbf{Mean (\%)} & \textbf{Std (\%)} & \textbf{Cost (vLLM)} & \textbf{Cost (z.ai)} \\
\midrule
400 & 34.40 & 33.00 & 33.00 & 33.47 & 0.81 & \$75 & \$37 \\
750 & 36.80 & 35.00 & 37.40 & 36.40 & 1.25 & \$140 & \$69 \\
1,500 & 38.20 & 40.20 & 38.20 & 38.87 & 1.15 & \$280 & \$138 \\
3,000 & 40.60 & 37.80 & 40.60 & 39.67 & 1.62 & \$560 & \$275 \\
4,200 & 40.60 & 45.80 & 39.00 & 41.80 & 3.56 & \$784 & \$386 \\
7,400 & 43.20 & 45.40 & 43.40 & 44.00 & 1.22 & \$1,382 & \$679 \\
16,000 & 47.00 & 47.00 & 45.80 & 46.60 & 0.69 & \$2,987 & \$1,469 \\
\bottomrule
\end{tabular}
\end{table}

\section{Additional Baseline Comparisons}
\label{sec:additional_baselines}

Table~\ref{tab:additional-baselines} provides additional baseline comparisons that complement the main results in Table~\ref{tab:swebench-verified}. We train Qwen 2.5-32B  on SWE-smith, which performs much better than when transferred to Qwen 3. This suggests that SWE-smith is optimized for Qwen 2.5-32B. We also include the \sera result using GLM-4.5-Air as a teacher, which shows the substantial performance improvement from using a stronger teacher model compared to Claude 3.7.

\begin{table}[h!]
\centering
\caption{Additional baseline comparisons. SWE-smith with Qwen 2.5-32B shows the method was optimized for this model family. \sera with GLM-4.5-Air demonstrates the benefit of stronger teacher models.}
\label{tab:additional-baselines}
\begin{tabular}{l|c|c}
\toprule
Method & SWE-smith & \sera \\
\midrule
Base model & Qwen 2.5-32B & Qwen 3-32B \\
Teacher & Claude 3.7 & GLM-4.5-Air \\
Eval context size & 32K & 32K \\
Sample size & 6402 & 4933 \\
\midrule
SWE-bench Verified & 32.60\% & 38.47\% $\pm$ 1.01\% \\
\bottomrule
\end{tabular}
\end{table}

\section{Specialization Results at 64K Context}
\label{sec:specialization-64k}

Table~\ref{tab:specialization-64k} presents specialization results evaluated at 64K context length. Because \sera models are trained at 32K context while competing models like Devstral-Small-2 are trained at 64K or longer contexts, \sera underperforms at 64K evaluation despite matching or exceeding these baselines at 32K (Table~\ref{tab:specialization}). This context length mismatch explains the performance gap: our models have not learned to effectively utilize the additional context available at 64K tokens.

\begin{table}[h!]
\centering
\caption{Specialization results at 64K context. Fine-tuned \sera models underperform baselines at 64K because they are trained at 32K context, while Devstral-Small-2 is trained at longer contexts. Results averaged over three seeds.}
\label{tab:specialization-64k}
\begin{tabular}{l|ccc}
\toprule
    {\textbf{\small{Model}}} &
    {\textbf{\small{Django (231)}}} &
    {\textbf{\small{Sympy (75)}}} &
    {\textbf{\small{Sphinx (44)}}}
\\
\midrule
{Qwen 3-32B-Django} & 56.56\% $\pm$ 0.66\% & - & - \\
{Qwen 3-32B-Sympy} & - & 48.00\% $\pm$ 4.62\% & - \\
{Qwen 3-32B-Sphinx} & - & - & 35.61\% $\pm$ 1.31\% \\
\midrule
\rowcolor{gray!20}{GLM-4.5-Air} & 58.58\% $\pm$ 1.39\% & 56.00\% $\pm$ 1.33\% & 48.87\% $\pm$ 1.98\% \\
\rowcolor{gray!20}{Devstral-Small-2-24B} & \textbf{62.63\% $\pm$ 1.32\%} & \textbf{56.24\% $\pm$ 3.27\%} & \textbf{53.79\% $\pm$ 4.73\%} \\
\bottomrule
\end{tabular}
\end{table}

\section{Cost Breakdown}
\label{sec:cost}

We assume a cost of \$2 per H100 GPU-hour throughout this section, which reflects current cloud pricing for on-demand instances.

\textbf{Reinforcement Learning:} RL-based approaches for coding agents require substantial compute. SkyRL-Agent \citep{cao2025skyrl} reports 4,601 H100-hours to train SA-SWE-32B, yielding a cost of \$9,202 and achieving 39.4\% on SWE-bench Verified. For comparison, DeepSWE \citep{luo2025deepswe} requires 9,180 H100-hours (\$18,360) to reach similar performance. \sera's total cost for data generation and training is 960 H100-hours (\$1,920) to match DeepSWE's performance, 9.6$\times$ cheaper than DeepSWE. \sera also achieves higher data efficiency. We fit a power law to \sera's cost-performance curve (Figure~\ref{fig:scaling}) and find that \sera reaches SkyRL's 39.4\% at a cost of just \$352 when self-hosting via vLLM, or \$173 via the z.ai API. This yields a cost-to-performance efficiency of 26$\times$ (vLLM) or 53$\times$ (z.ai) compared to SkyRL.

\textbf{Synthetic Data Generation:} Figure~\ref{fig:scaling} shows scaling curves under three cost regimes: self-hosted inference via vLLM, and API-based inference using GLM-4.5-Air and GLM-4.6 through the z.ai API. To derive the API cost, we analyzed 100 randomly sampled trajectories from the SWE-smith trajectory dataset \citep{yang2025swesmith} to measure actual token consumption patterns. Each trajectory consists of multiple API calls where the conversation history grows with each turn. For a given API call, the model receives the full conversation history (cached input), the new tool result or observation (uncached input), and produces a response (output). We measured these components and rescaled to 32K context length to match our training setup, yielding an average of 35 API calls per trajectory.

Table~\ref{tab:cost_breakdown} shows the per-trajectory cost breakdown across four configurations: SWE-smith using the Sonnet 3.7 API, \sera using the z.ai API with GLM-4.5-Air and GLM-4.6, and \sera self-hosted via vLLM. For the API-based methods, we show the token-level breakdown; for vLLM, we report the GPU cost directly.

\begin{table}[h!]
\centering
\caption{API pricing used for cost calculations.}
\label{tab:api_pricing}
\begin{tabular}{l|c|c|c}
\toprule
\textbf{Provider} & \textbf{Input (/MTok)} & \textbf{Cached (/MTok)} & \textbf{Output (/MTok)} \\
\midrule
Anthropic (Sonnet 3.7) & \$3.00 & \$0.30 & \$15.00 \\
z.ai (GLM-4.5-Air) & \$0.20 & \$0.03 & \$1.10 \\
z.ai (GLM-4.6) & \$0.60 & \$0.11 & \$2.20 \\
vLLM (self-hosted) & \multicolumn{3}{c}{0.065 GPU-hours/trajectory $\times$ \$2/GPU-hour} \\
\bottomrule
\end{tabular}
\end{table}

\begin{table}[h!]
\centering
\caption{Cost breakdown per trajectory. Token cost percentages show the share of total billed tokens across 35 API calls per trajectory, rescaled to 32K context. See Table~\ref{tab:api_pricing} for pricing details.}
\label{tab:cost_breakdown}
\begin{tabular}{l|r||r||r||r||r}
\toprule
& & \textbf{SWE-smith} & \textbf{\sera} & \textbf{\sera} & \textbf{\sera} \\
\textbf{Component} & \textbf{\% Token Cost} & \textbf{(Sonnet 3.7)} & \textbf{(GLM-4.5-Air)} & \textbf{(GLM-4.6)} & \textbf{(vLLM)} \\
\midrule
Cached input (context) & 95.9\% & \$0.2247 & \$0.0225 & \$0.0824 & --- \\
New input (tool results) & 3.1\% & \$0.0730 & \$0.0049 & \$0.0146 & --- \\
Output (generations) & 1.0\% & \$0.1151 & \$0.0084 & \$0.0169 & --- \\
Issue creation & --- & \$0.0540 & --- & --- & --- \\
\midrule
\textbf{Inference subtotal} & --- & \$0.4668 & \$0.0358 & \$0.1139 & \$0.1307 \\
Training & --- & \$0.0560 & \$0.0560 & \$0.0560 & \$0.0560 \\
\midrule
\textbf{Total per trajectory} & --- & \textbf{\$0.5228} & \textbf{\$0.0918} & \textbf{\$0.1699} & \textbf{\$0.1867} \\
\bottomrule
\end{tabular}
\end{table}

The dominant cost for API-based methods is the cached conversation context, which accumulates approximately 749K tokens across 35 API calls per trajectory. For Sonnet 3.7, even with prompt caching at \$0.30/MTok, the cumulative context accounts for 54.4\% of inference cost. Output tokens, though far fewer (7.7K per trajectory), are disproportionately expensive due to the higher output price (\$15.00/MTok). SWE-smith additionally requires \$0.054 per trajectory for synthetic issue creation. In total, \sera with GLM-4.5-Air via the z.ai API is 5.7$\times$ cheaper than SWE-smith with Sonnet 3.7, and 2.0$\times$ cheaper than self-hosting via vLLM. GLM-4.6 via the z.ai API costs \$0.1699 per trajectory---3.1$\times$ cheaper than SWE-smith and 1.1$\times$ cheaper than vLLM self-hosting, while providing a stronger teacher model. The 3.2$\times$ higher inference cost of GLM-4.6 compared to GLM-4.5-Air (\$0.1139 vs \$0.0358) is partially offset by the fixed training cost (\$0.056), yielding only a 1.85$\times$ increase in total per-trajectory cost. Importantly, these per-trajectory comparisons do not account for data quality. As shown in Table~\ref{tab:swebench-verified}, \sera achieves higher performance per sample than competing methods, and GLM-4.6 produces higher-quality data than GLM-4.5-Air at comparable sample sizes. When we account for this by comparing the cost to reach equivalent performance levels using our scaling law, the effective advantages are substantially larger: \sera reaches SWE-smith's 32.6\% (Qwen 2.5) performance at a cost of \$60 (vLLM) or \$29 (z.ai with GLM-4.5-Air), compared to SWE-smith's \$3,395. This yields a cost-to-performance efficiency of 57$\times$ (vLLM) or 115$\times$ (z.ai with GLM-4.5-Air).

At scale, \sera requires approximately \$1.5K to generate 16,000 trajectories via the z.ai API with GLM-4.5-Air, \$2.7K with GLM-4.6, compared to \$3.0K via vLLM and \$8.4K via the Sonnet 3.7 API. The scaling law in Figure~\ref{fig:scaling} predicts that with GLM-4.6 via the z.ai API, matching Devstral-Small-2 performance requires approximately \$6K in data generation cost, compared to \$23K with GLM-4.5-Air via the z.ai API and \$47K with vLLM self-hosting.

However, we note important caveats for using commercial APIs in research. API pricing is subject to change, and providers may adjust model quality, rate limits, or availability without notice. This makes experiments difficult to reproduce exactly and can introduce confounding factors if model behavior shifts between experimental runs. For these reasons, APIs may not be suitable for rigorous scientific work that demands full reproducibility. We still encourage researchers to consider the vLLM backend with open-weight models, which provides complete control over the inference process and ensures consistent behavior across experimental runs. That said, for practitioners operating under cost constraints who need to generate training data quickly, a commercial API with cached input pricing offers a viable alternative at substantially reduced cost.
